\documentclass{article}
\pdfoutput=1
\usepackage{microtype}
\usepackage{graphicx}
\usepackage{subfigure}
\usepackage{booktabs} 
\usepackage{multirow}
\usepackage{diagbox}
\usepackage{subfigure}
\usepackage{longtable}
\usepackage{listings}
\usepackage{color}
\definecolor{codegreen}{rgb}{0,0.6,0}
\definecolor{codegray}{rgb}{0.5,0.5,0.5}
\definecolor{codepurple}{rgb}{0.58,0,0.82}
\definecolor{backcolour}{rgb}{0.95,0.95,0.92}

\lstdefinestyle{mystyle}{
    backgroundcolor=\color{backcolour},   
    commentstyle=\color{codegreen},
    keywordstyle=\color{magenta},
    numberstyle=\tiny\color{codegray},
    stringstyle=\color{codepurple},
    basicstyle=\ttfamily\footnotesize,
    breakatwhitespace=false,         
    breaklines=true,                 
    captionpos=b,                    
    keepspaces=true,                 
    numbers=left,                    
    numbersep=5pt,                  
    showspaces=false,                
    showstringspaces=false,
    showtabs=false,                  
    tabsize=2
}

\usepackage{hyperref}

\usepackage[accepted]{icml2024}

\usepackage{amsmath}
\usepackage{amssymb}
\usepackage{mathtools}
\usepackage{amsthm}

\usepackage[capitalize,noabbrev]{cleveref}

\theoremstyle{plain}

\theoremstyle{definition}

\theoremstyle{remark}

\usepackage[textsize=tiny]{todonotes}

\icmltitlerunning{Learning Solution-Aware Transformers for Efficiently Solving Quadratic Assignment Problem}

\begin{document}

\twocolumn[

\icmltitle{Learning Solution-Aware Transformers for Efficiently Solving Quadratic Assignment Problem}
           


\icmlsetsymbol{equal}{*}

\begin{icmlauthorlist}
\icmlauthor{Zhentao Tan}{yyyy,yyy}
\icmlauthor{Yadong Mu}{yyy}
\end{icmlauthorlist}

\icmlaffiliation{yyyy}{Center for Data Science, Peking University, China.}
\icmlaffiliation{yyy}{Wangxuan Institute of Computer Technology, Peking University, China}

\icmlcorrespondingauthor{Yadong Mu}{myd@pku.edu.cn}


\vskip 0.3in
]



\printAffiliationsAndNotice{}  

\begin{abstract}
Recently various optimization problems, such as Mixed Integer Linear Programming Problems (MILPs), have undergone comprehensive investigation, leveraging the capabilities of machine learning. This work focuses on learning-based solutions for efficiently solving the Quadratic Assignment Problem (QAPs), which stands as a formidable challenge in combinatorial optimization.
While many instances of simpler problems admit fully polynomial-time approximate solution (FPTAS), QAP is shown to be strongly NP-hard. Even finding a FPTAS for QAP is difficult, in the sense that the existence of a FPTAS implies $P = NP$. Current research on QAPs suffer from limited scale and computational inefficiency. To attack the aforementioned issues, we here propose the first solution of its kind for QAP in the learn-to-improve category. This work encodes facility and location nodes separately, instead of forming computationally intensive association graphs prevalent in current approaches. This design choice enables scalability to larger problem sizes. Furthermore, a \textbf{S}olution \textbf{AW}are \textbf{T}ransformer (SAWT) architecture integrates the incumbent solution matrix with the attention score to effectively capture higher-order information of the QAPs. Our model's effectiveness  is validated through extensive experiments on self-generated QAP instances of varying sizes and the QAPLIB benchmark.


\end{abstract}
\section{Introduction}

Combinatorial Optimization Problems (COPs) involving discrete variables as input have garnered substantial attention in diverse real-world applications. This includes in-depth studies related to vehicle routing problems (VRPs) \cite{DBLP:journals/tits/VeresM20}, chip placement and routing problems \cite{DBLP:journals/nature/MirhoseiniGYJSW21,DBLP:conf/nips/LaiM022,tan2024hierarchical}, and molecule learning \cite{DBLP:conf/nips/AhnKLS20,DBLP:conf/www/YangZWY23}. Given the NP-hard nature of COPs, conventional solvers and manually crafted heuristics face challenges in finding optimal solutions. Consequently, there has been a recent emergence of machine learning based methods aimed at swiftly identifying sub-optimal solutions, proving notably successful in this pursuit.


Given the discrete nature of COPs, reinforcement learning (RL) proves instrumental in their resolution, either by constructing solutions in learn-to-construct (L2C) methods~\cite{vinyals2015pointer,kwon2020pomo} or improving initial solutions iteratively in learn-to-improve (L2I) solvers~\cite{ma2021learning,ma2023learning}. Despite the prevalent focus on mixed integer linear programming problems (MILPs), particularly in solving VRPs~\cite{DBLP:journals/ijon/ZhangLLZYLY23}, machine learning applications to QAPs remain relatively sparse in current literature.


For QAPs, prevalent machine learning approaches focus on graph matching~\cite{nowak2018revised,wang2021neural,liu2022revocable}. Specifically, \cite{wang2021neural} utilizes an association graph to address QAPs with a complexity of $O(n^4)$ for instances of size $n$, limiting its applicability to only small problem size (\emph{e.g.}, below 30). Moreover, it requires labeled data for training, impractical for larger QAP sizes. Following~\cite{wang2021neural}, the work in~\cite{liu2022revocable} employs L2C reinforcement learning methods for QAPs without requiring ground truth, yet scalability challenges persist. In contrast, traditional Operations Research (OR) approaches \cite{zhang2020hybrid,mihic2018randomized} rely on meta-heuristic search strategies. For instance, \cite{zhang2020hybrid} combines genetic heuristics with tabu-search, and \cite{mihic2018randomized} adopts an adaptive large neighbor search strategy. However, these methods often demand hours to find the optimal solution for a single instance, rendering them inefficient.


To address the outlined challenges, we propose a novel L2I approach that integrates reinforcement learning with a Solution Aware Transformer (SAWT) model to tackle Koopmans-Beckmann's Quadratic Assignment Problem. This method adeptly handles QAP instances of various sizes. To eliminate the need for an association graph, we employ a mixed-score Transformer and a Graph Convolution Network (GCN) to independently encode facility and location nodes, enabling scalability to larger problem sizes. Our SAWT encoder dynamically captures graph structure patterns in the QAP, adapting to different assignment solutions. The decoder facilitates refinement by executing swap operations between two positions within the assignment solution, enhancing overall quality. Extensive experiments on self-generated QAP instances and~QAPLIB \cite{burkard1997qaplib} benchmarks showcase the model's effectiveness and robust generalization to instances of varying sizes. Most importantly, to our best knowledge, the proposed method is the first L2I method of its kind to solve the QAP.


The paper makes the following key contributions: (1) Introducing the first learn-to-improve reinforcement learning method for the Quadratic Assignment Problem, exhibiting proficiency in solving instances up to a size of 100. (2) Proposing a novel Solution Aware Transformer (SAWT) model adept at effectively capturing QAP patterns through the dynamic integration of solution-aware information into the attention model. (3) Through extensive experiments on self-generated Koopmans-Beckmann's QAP instances and QAPLIB benchmarks, showing the SAWT model's efficient QAP-solving capabilities and robust generalization across various instances.


\section{Quadratic Assignment Problem}
\label{QAP_section}

As per the definition in~\cite{koopmans1957assignment}, a QAP involves the optimal assignment of $n$ facilities to $n$ locations. Represented with a facility set $\mathbf{Fac}$ and a location set $\mathbf{Loc}$, the Koopmans-Beckmann's QAP is defined as below:
\begin{equation}
\renewcommand{\arraystretch}{1.5}
\setlength{\arraycolsep}{2pt}
\begin{array}{ll}
\mathop{\min}\limits_{x} & \displaystyle\sum_{i, j=1}^n \sum_{k, p=1}^n f_{i j} d_{k p} x_{i k} x_{j p}, \\
\text{s.t.} & \displaystyle\sum_{i=1}^n x_{i j}=1, \ 1 \leqslant j \leqslant n, \\
& \displaystyle\sum_{j=1}^n x_{i j}=1, \ 1 \leqslant i \leqslant n, \\
& x_{i j} \in\{0,1\}, \ 1 \leqslant i, j \leqslant n ,
\end{array}
\label{QAP}
\end{equation}
where $f_{ij}$ is the flow from facility $fac_i$ to facility $fac_j$, and $d_{kp}$ is the distance between locations $loc_k$ and location $loc_p$. $x_{ik}$ is a binary variable indicating whether to place facility $fac_i$ to location $loc_k$ or not. The constraints in Definition~\ref{QAP} stipulate the exclusive assignment of one facility to a single location. Another prevalent QAP formulation is presented in trace format~\cite{edwards1980branch}:
\begin{equation}
\label{trace QAP}
\renewcommand{\arraystretch}{1.5}
\setlength{\arraycolsep}{2pt}
\begin{array}{ll}
    \mathop{\min}\limits_{\mathbf{X}} & \displaystyle trace (\mathbf{F} \cdot \mathbf{X} \cdot \mathbf{D} \cdot \mathbf{X}^{T}), \\
\text{s.t.} & \displaystyle \mathbf{X} \cdot \mathbf{1} = \mathbf{1}, \mathbf{X}^{T} \cdot \mathbf{1} = \mathbf{1}, \\
& \mathbf{X} \in\{0,1\} ^{n \times n} ,
\end{array}
\end{equation}
where $\mathbf{F}$ and $\mathbf{D}$ are flow or distance matrix, respectively. $\mathbf{X}$ is a permutation matrix and $\mathbf{1}$ is an all-one vector. For clarity, we set the objective function for the QAP as $L(\mathbf{X}) = trace (\mathbf{F} \cdot \mathbf{X} \cdot \mathbf{D} \cdot \mathbf{X}^{T})$. 

The QAP finds relevance in diverse real-world applications. For instance, it has  applications in electronic module placement~\cite{53bbb073-5adc-3787-97ba-b0c9dae99c60}, where $f_{ij}$ denotes connections between modules $i$ and $j$, and $d_{kp}$ signifies distances between locations $k$ and $p$, allowing for the minimization of total electrical connection length. Additionally, QAP was also utilized in hospital room assignment scenarios~\cite{elshafei1977hospital}, where $f_{ij}$ represents patient transfers between rooms, and $d_{ij}$ indicates travel time from room $i$ to room $j$. Other applications include imagery~\cite{taillard1991robust} and turbine runner balancing~\cite{LAPORTE1988378}. For further details, interested readers can refer to the provided references.


Critically, solving the QAP is challenging, as evidenced by~\cite{DBLP:journals/jacm/SahniG76}, which establishes the absence of a polynomial algorithm with $\epsilon$-approximation unless $P = NP$. Notably, QAP, encompassing the well-studied Traveling Salesman Problem (TSP) as a special case, emerges as a more intricate problem. Despite its widespread application, the current deep learning landscape has shown limited attention to QAP. Considering the real-world significance and inherent complexities, we posit that deep learning methods can play a substantial role in addressing and influencing solutions for QAPs.


\section{Related Work}


\textbf{Traditional methods for QAP}: Koopmans-Beckmann's QAP has received significant attention in the Operations Research (OR) community. Genetic algorithms~\cite{hanh2019efficient,ahmed2015multi} were probability search methods based on biological principles of natural selection, recombination, mutation, and survival of the fittest. Tabu-search algorithms~\cite{zhang2020hybrid,shylo2017solving,james2009cooperative} were local search methods that utilize a tabu list to prevent duplicate solutions. Other meta-heuristics like large neighborhood search~\cite{mihic2018randomized,wang2023new} and swarm algorithms~\cite{cui2023uav} show competitive performance on QAPs. However, these methods, tailored to specific problems, often require hours for solving a single instance, posing challenges for real-world deployment.


\textbf{Learning-based methods for QAP}: Learning-based methods for QAPs remain limited in the literature. The majority of existing works focus on graph matching~\cite{DBLP:conf/cvpr/ZanfirS18,yu2021deep,lin2023graph}. The work in~\cite{wang2021neural} pioneered supervised learning for QAPs, transforming them into node classification tasks on association graphs, where each node represents a matching pair. Classification of a node as one indicates the selection of that pair for matching. The state-of-the-art work in~\cite{liu2022revocable} introduced the first learn-to-construct (L2C) reinforcement learning method for QAPs without ground truth, utilizing struct2vec~\cite{dai2016discriminative} for node encoding. However, scalability issues arise due to the $O(n^4)$ complexity of association graphs.


\textbf{RL for combinatorial optimization}: Reinforcement learning for COPs has witnessed extensive recent research. Learn-to-construct (L2C) methods began with~\cite{DBLP:conf/iclr/BelloPL0B17}, utilizing PtrNet~\cite{vinyals2015pointer} for solving Traveling Salesman Problems. Subsequent advancements involved Graph Neural Networks (GNNs)~\cite{khalil2017learning}, attention models~\cite{kwon2020pomo} for high-quality solutions in vehicle routing problems, Minimum Vertex Cover, and other COPs. MatNet~\cite{DBLP:conf/nips/KwonCYPPG21} is the first to take matrix inputs for COP resolution. In learn-to-improve (L2I) methods, \cite{chen2019learning} relied on local search, improved by NLNS solver~\cite{wu2021learning} with handcrafted operators. Recent approaches focus on controlling k-opt and swap heuristics for VRPs, with~\cite{d2020learning} on 2-opt, \cite{sui2021learning}~on 3-opt, and~\cite{ma2021learning,ma2023learning} achieving state-of-the-art performance using Transformer and special positional embeddings to capture the linear structure of VRPs. However, these methods lack the capability to address QAPs due to their inability to capture the graph structure within QAPs.


\section{Problem Formulation}

As explained in Section \ref{QAP_section}, the solution to a QAP involves creating a one-to-one assignment between facilities and locations that minimizes the total cost. An assignment: $\sigma = (\sigma(1), ..., \sigma(n))$ is a map where $\sigma : i \rightarrow \sigma(i)$ means mapping facility $fac_i$ to location $loc_{\sigma(i)}$. 

Starting with an initial feasible solution, our deep reinforcement learning model iteratively enhances the solution. The policy initiates by selecting a facility pair $(i, j; i < j)$, and subsequently, a swap operation is applied to the current solution $\sigma$, resulting in the next solution $\sigma' = (\sigma(1), ..., \sigma(j), ..., \sigma(i), ..., \sigma(n))$. This swap operation iterates until reaching the step limit $T_{limit}$, encapsulating the process as a Markov Decision Process (MDP), as depicted below.


\textbf{States.} Following \cite{d2020learning}, we define a state $\Bar{\sigma} = (\sigma, \sigma^{\ast})$ where $\sigma$ and $\sigma^{\ast}$ are the current solution and current lowest-cost solution respectively. For example, at step $t$, $\sigma_t^{\ast} = \arg \min_{\sigma_{\Bar{t}} \in \{\sigma_1,...,\sigma_t\}} L(\sigma_{\Bar{t}}) $.

\textbf{Actions.} The actions in our model are pairs of facilities' indices: $A = (i,j; ~~i < j) $ for the swap operations.

\textbf{Rewards.} The reward function is defined as: $r_t = L(\sigma_{t}^{\ast}) - \min (L(\sigma_{t}^{\ast}), L(\sigma_{t+1}))$. This formulation ensures that the sum of intermediate rewards at each step corresponds to the total decrease in cost relative to the initial solution.

\textbf{Transitions.} Given the action: $(i,j)$ and current solution $\sigma_t$, the next solution is generated through swap operation: $\sigma_{t+1} = (\sigma_t(1), ..., \sigma_t(j), ..., \sigma_t(i), ..., \sigma_t(n))$. Note that even if the transition does not improve the objective function, the RL agent will still accept the next solution. By doing this, we can expect the agent to punish non-improving actions by assigning 0 rewards. 


\textbf{Policy.} The policy $\pi_\theta$ is parameterized by our SAWT model with the parameter $\theta$. At step $t$, given the current state $\sigma_t$, an action pair $a_t = (i_t, j_t)$ is sampled through $\pi_\theta(a_t | \sigma_t)$. For an episode with a length of $T_{limit}$ starting from the initial solution $\sigma_0$, the process can be expressed as the probability based on the chain rule:
\begin{equation}
    P(\sigma_{T_{limit}} | \sigma_0) = \prod_{t=1}^{T_{limit}} \pi_\theta(a_t | \sigma_{t-1}).
\end{equation}

\begin{figure*}[t]
    \centering
    \includegraphics[width=0.95\textwidth]{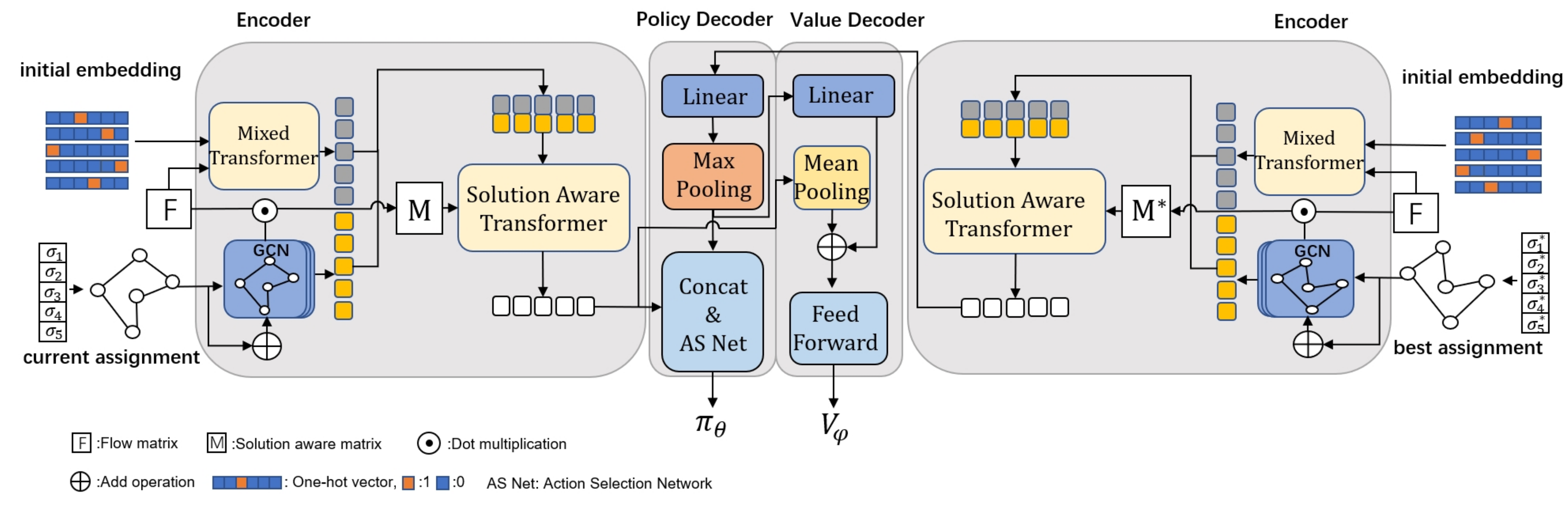}
    \caption{Architecture of our policy network, namely Solution Aware Transformer (SAWT). See main text for more details.}
    \label{model architecture}
\end{figure*}

\section{Solution-Aware Transformers}

\begin{figure}[t]
    \centering
    \includegraphics[width=0.9\linewidth]{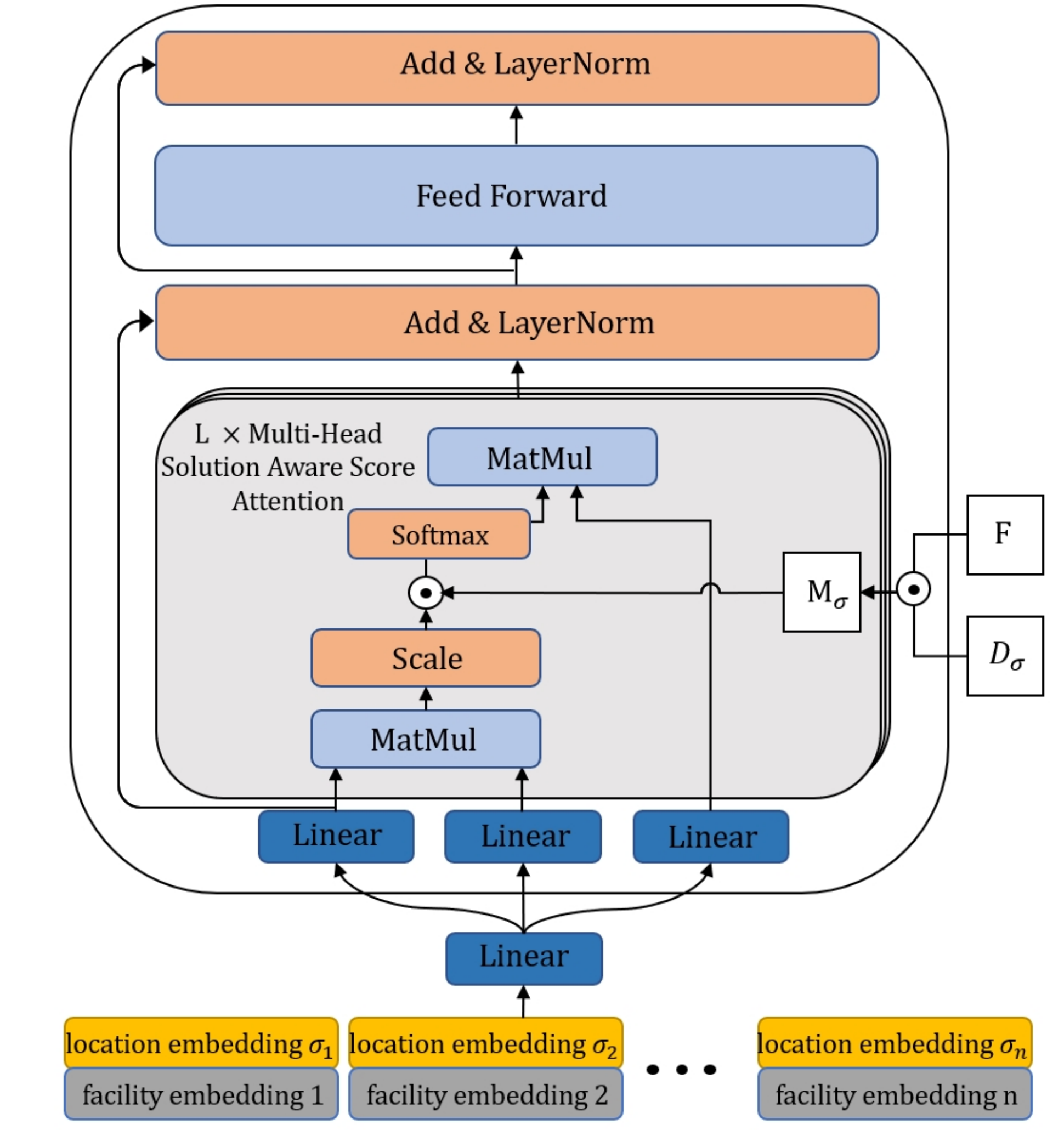}
    \caption{Architecture of SAWT module. For a given current solution $\sigma$, we rearrange the rows of the distance matrix to create $\mathbf{D}\sigma = (\mathbf{d}{\sigma(1)}, ..., \mathbf{d}{\sigma(n)})$. Conducting a dot product between the flow matrix $\mathbf{F}$ and the distance matrix $\mathbf{D_\sigma}$ results in the solution-aware matrix $\mathbf{M}\sigma$, which encapsulates the complete information about the QAP's objective. This is because the cost for the current solution $\sigma$ is the sum of $\mathbf{M}\sigma$. By integrating $\mathbf{M}_\sigma$ into the attention module, our goal is to equip the model with incumbent solution information.}
    \label{SAWT module}
\end{figure}

Here we introduce our policy network: Solution Aware Transformer (SAWT). The entire model pipeline is depicted in Figure~\ref{model architecture}, illustrating the process using a QAP example with 5 facilities and locations. In the SAWT \textit{encoder} module, the SAWT processes the concatenation of embeddings for 5 facilities and locations, incorporating the solution-aware matrix $\mathbf{M}$ into the attention module. This design enables our model to dynamically capture information from the incumbent solution. The SAWT \textit{decoder} then utilizes the output from the \textit{encoder} to generate action pairs and critic values.


\subsection{Embeddings}
\label{embeddings}


To circumvent the need for an association graph, we independently encode facility and location nodes, improving scalability to larger problem sizes.



\paragraph{Location embedding.} The initial representation of a location, $loc_i = (x_i, y_i)$, is a 2-dimensional coordinate. We employ a linear projection function on $loc_i$ to obtain a latent representation $loc_i \in \mathbb{R}^{d}$. Locations form a distance graph so that it is natural to encode them using Graph Convolutional Network (GCN). $loc_i$ is then fed into three layers of a GCN, as defined in Equation~\ref{GCN}, using the distance matrix as the affinity matrix, resulting in the final embedding of locations $loc_i \in \mathbb{R}^{d}$.

\begin{equation}
\label{GCN}
    \mathbf{Loc}^{(l+1)} = \mathbf{Loc}^{(l)} + \text{ReLU}(\mathbf{D} ~ \mathbf{Loc}^{(l)} ~ \mathbf{W}^{(l)}),
\end{equation}

where $\mathbf{W}^{(l)} \in  \mathbb{R}^{d \times d}$ are the trainable parameters and ReLU refers to the Rectified Linear Unit.


\paragraph{Facility embedding.} Encoding facility nodes into a higher-dimensional space poses a challenge due to the matrix-type input (flow matrix), which neural networks find difficult to encode. Drawing inspiration from~\cite{DBLP:conf/nips/KwonCYPPG21}, we address this by mixing the flow matrix into the attention score within a Transformer. Initialization of facility nodes' embeddings $fac_i$ begins with a random one-hot vector of a predefined high dimension $N_{init} >> n$. The encoding process for facility nodes then follows the equation as shown below:
\begin{equation}
\begin{gathered}
\mathbf{Fac}^{(l+1)}  = \text{softmax}(g(\textbf{Att}(\mathbf{Fac}^{(l)},  \mathbf{Fac}^{(l)}) , \mathbf{F})) (\mathbf{Fac}^{(l)}\mathbf{W}^v), \\
\textbf{Att}(\mathbf{Fac}^{(l)},\mathbf{Fac}^{(l)}) = \frac{(\mathbf{Fac}^{(l)}\mathbf{W}^q) (\mathbf{Fac}^{(l)}\mathbf{W}^k)} {\sqrt{N_{init}}},
\end{gathered}
\end{equation}
where $\mathbf{W}^q \in \mathbb{R}^{N_{init} \times d_q}, \mathbf{W}^k \in \mathbb{R}^{N_{init} \times d_k}, \mathbf{W}^v \in \mathbb{R}^{N_{init} \times d_v}$ are learnable parameters. $\mathbf{F}$ and $\mathbf{Fac}^{(0)}$ are the flow matrix and one-hot initial embedding, respectively. $g$ is a linear function that takes $\mathbf{F}$ and $\textbf{Att}(\mathbf{Fac}^{(l)},\mathbf{Fac}^{(l)})$ as input. Practically, we find our model works best for $g$ equals to dot product operation.


Our embedding strategy diverges from~\cite{DBLP:conf/nips/KwonCYPPG21} by intentionally restricting early interaction between facility and location nodes within the attention layer. This distinction arises from our consideration of facility nodes as separate entities from location nodes. Additionally, recognizing that a flow matrix, when paired with different distance matrices, always forms a valid QAP, our focus is on independently learning facility node embeddings and location nodes embeddings without involving interaction. Experimental results later substantiate the effectiveness of our embedding strategy.

\subsection{The Encoder}


The encoder comprises $L$ stacked SAWT encoders, and the architecture of a SAWT encoder is illustrated in Figure~\ref{SAWT module}. It takes the concatenation of facility embeddings $\{fac_i \}_{i=1}^n$ and location embeddings $\{loc_i \}_{i = 1}^n$ as input, passing through a Multi-Head Solution Aware Attention (\textbf{SAWT-Att}) sub-layer and a feed-forward network sub-layer. Each sub-layer is succeeded by skip connection~\cite{DBLP:conf/cvpr/HeZRS16} and layer normalization~\cite{DBLP:journals/corr/BaKH16}, mirroring the original Transformer design. The encoding process is shown in Equation~\ref{SAWT encoder}.
\begin{equation}
\label{SAWT encoder}
\begin{gathered}
    h^{'}_i = \textbf{LN}(h_i^{(l)} + \textbf{SAWT}(\mathbf{H}^{(l)},\mathbf{F},\mathbf{D}_\sigma)_i), \\ 
    h_i^{(l+1)} = \textbf{LN}(h^{'}_i + \textbf{FFN}(h^{'}_i)).
\end{gathered}
\end{equation}
\paragraph{SAWT-Att.} The SAWT-Att aims to utilize the current solution information to enhance the embeddings. SAWT-Att takes the concatenation of facility embeddings and location embeddings, denoted as $\mathbf{H}^{(0)}$ as input. This concatenation is performed based on the current solution $\sigma$.
\begin{equation}
\label{initial input}
    \mathbf{H}^{(0)} = [ \mathbf{Fac} || \mathbf{Loc}_\sigma ] \mathbf{W}^{(0)}, \mathbf{Loc}_\sigma = \mathbf{X}_\sigma \mathbf{Loc},
\end{equation}
where $\cdot||\cdot$ is the concatenation operation, $\mathbf{W}^{(0)} \in \mathbb{R}^{2d \times d}$ is a trainable parameter matrix and $\mathbf{X}_\sigma$ is the permutation matrix where $\mathbf{X}_{i,\sigma(i)} = 1$ and other elements equal to 0. Following Equation~\ref{initial input}, $\mathbf{Loc}_\sigma = (\mathbf{loc}_{\sigma(1)}, ..., \mathbf{loc}_{\sigma(n)})$. Based on the definition that $\sigma(i)$ assigns $fac_i$ to $loc_{\sigma(i)}$, it is reasonable to concatenate the facility nodes and location nodes in the way presented in Equation~\ref{initial input}. Then we compute the solution aware self-attention correlation following the rule below:
\begin{equation}
\label{SAWT-Att}
\begin{gathered}
    \textbf{SAWT-Att}(\mathbf{H}^{(l)},\mathbf{F},\mathbf{D}_\sigma)  = \textbf{Att}(\mathbf{H}^{(l)},\mathbf{H}^{(l)}) \odot \mathbf{M}_\sigma, \\
    \mathbf{M}_\sigma = \mathbf{F} \odot \mathbf{D}_\sigma, \mathbf{D}_\sigma = \mathbf{X}_\sigma \mathbf{D}, 
\end{gathered}
\end{equation}
where $\odot$ is element-wise multiplication. 


The self-attention module $\textbf{Att}(\mathbf{H}^{(l)}, \mathbf{H}^{(l)})$ captures the node-wise relationships within $\mathbf{H}^{(l)}$. The matrix $\mathbf{M}_\sigma$, on the other hand, encapsulates the edge-wise relationships present in $\mathbf{H}^{(l)}$. This is evident as $\mathbf{M}_{\sigma(i,j)} = \mathbf{F}_{i,j} \cdot \mathbf{D}_{\sigma(i),\sigma(j)}$ represents the cost associated with assigning facilities $i$ and $j$ to locations $\sigma(i)$ and $\sigma(j)$ respectively. Crucially, $\mathbf{M}_\sigma$ is solution-aware, given that $\text{Sum}( \mathbf{M}_\sigma) = trace (\mathbf{F} \cdot \mathbf{X}_\sigma \cdot \mathbf{D} \cdot \mathbf{X}_\sigma^{T})$ from Equation~\ref{trace QAP} denotes the QAP cost with the solution $\sigma$. By combining $\mathbf{M}_\sigma$ with self-attention correlation $\textbf{Att}(\mathbf{H}^{(l)}, \mathbf{H}^{(l)})$, we aim to enhance the model's ability to capture diverse QAP patterns with different solutions. Importantly, SAWT incorporates the objective gradient information of QAP into its model design for the fact that $\mathbf{M}_\sigma \propto D_{X} trace(\mathbf{F} \cdot \mathbf{X} \cdot \mathbf{D} \cdot \mathbf{X}^{T})$ where details are in Appendix \ref{theory}. This allows $\mathbf{M}_\sigma$ to guide the model in the descent direction, leading to improved solutions. Subsequently, the output of \textbf{SAWT} is obtained using a multi-head attention scheme~\cite{DBLP:conf/nips/VaswaniSPUJGKP17}:
\begin{equation}
\begin{gathered}
    \textbf{SAWT}(\mathbf{H}^{(l)},\mathbf{F},\mathbf{D}_\sigma)_i = \text{Concat}[ \text{head}_{i,1}, ..., \text{head}_{i,m}] \mathbf{W}^O, \\
    \text{head}_{i,j} = (\text{softmax}(\textbf{SAWT-Att}(\mathbf{H}^{(l)},\mathbf{F},\mathbf{D}_\sigma)) (\mathbf{H}^{(l)} \mathbf{W}^{v}_j))_i,
\end{gathered}
\end{equation}
where $\mathbf{W}^v_j \in \mathbb{R}^{d \times d_v}$ and $\mathbf{W}^O \in \mathbb{R}^{md_v \times d}$ are trainable matrices. 

\begin{table*}[t]
\caption{The experiments were conducted on the QAP of sizes 10, 20, 50, and 100. For each size, we trained on 5120 instances with a batch size of 512 and evaluated 256 instances. ``MEAN" means the mean cost averaged on the total test instances. ``TIME" means the total time needed to solve the test instances. ``---" means that the method can not solve the QAP due to the limited CPU memory or GPU memory.  ``MatNet$^\ast$" indicates our adoption of the MatNet encoding method to implement an improvement strategy. ``$\ast\ast$" means that we use 4 CPU units to solve the test instances due to the long inference time. }
\centering
\tiny
\begin{sc}
\begin{tabular}{l|ccc|ccc|ccc|ccc}
\hline \multirow{2}{*}{method}  & \multicolumn{3}{|c|}{QAP10} & \multicolumn{3}{|c|}{QAP20} & \multicolumn{3}{|c|}{QAP50} & \multicolumn{3}{|c}{QAP100} \\
\cline{2-13} & Mean$\downarrow$ & Gap$\downarrow$  &  Time$\downarrow$  & Mean$\downarrow$ & Gap$\downarrow$  &  Time$\downarrow$ & Mean$\downarrow$ & Gap$\downarrow$  &  Time$\downarrow$  & Mean$\downarrow$ & Gap$\downarrow$  &  Time$\downarrow$  \\
\hline
Gurobi  & 11.79 & 0.00\% & 5m & 55.18 &0.45\% & 1d18h50m & 392.57& 3.27\%& 1d18h2m &---&---&--- \\
TS$\{1\text{k}\}$ & 11.84& 0.42\%& 2m3s& 55.16 &0.41\%&12m27s&380.56&0.11\%&2h27m&1615.77&0.43\%&12h32m$^{\ast\ast}$ \\
TS$\{5\text{k}\}$ & 11.79& 0.00\%&11m10s&54.93& 0.00\%&1h38m&380.13&0.00\%&17h56m&1608.82&0.00\%&6d23h34m$^{\ast\ast}$\\
\hline
SM& 16.29&38.16\%&\textbf{0.1s}&70.07&27.56\%&\textbf{1.1s}&446.62&17.49\%&\textbf{18.3s}&1800.75&11.92\%&\textbf{3m}\\
RRWM&17.28&46.56\%&6.7s&73.74&34.24\%&14.3s&457.40&20.32\%&39.6s&1823.76&13.36\%&4m \\
IPFP&17.55&48.85\%&0.4s&75.52&37.48\%&2.4s&479.44&26.12\%&47.4s&1911.90&18.83\%&5m\\
\hline
MatNet$^{\ast}$ $\{10\text{k}\}$&12.67&7.46\%&5m12s&61.89&12.67\%&17m29s&---&---&---&---&---&---\\
Costa $\{10\text{k}\}$ &12.08&2.45\%&2m15s&57.91&5.42\%&3m15s&404.28&6.35\%&5m34s&1720.62&6.94\%&9m20s\\
Sui $\{10\text{k}\}$ &12.04 &2.1\%&2m25s&56.75&3.32\%&3m58s&396.24&4.24\%&7m21s&1689.56&5.01\%&10m19s\\
Wu  $\{10\text{k}\}$ &12.36&4.83\%&2m27s&61.24&11.48\%&4m35s&425.06&11.81\%&9m45s&1766.88&9.82\%&13m37s\\
Dact  $\{10\text{k}\}$&12.30&4.32\%&1m59s &60.47&10.08\%& 2m38s&418.77&10.16\%&3m7s&1743.82&8.39\%&4m17s\\
Neuopt $\{10\text{k}\}$&12.46& 5.68\%& 3m33s&61.37&11.72\%&4m27s&430.61&13.27\%&9m23&1789.22&11.21\%&14m23s\\
\hline
RGM &12.00&1.78\%&51.2s &55.96& 1.87\%& 8m32s &---&---&---&---&---&---\\
\hline
SAWT$\{2\text{k}\}$ &11.87&0.67\%&15s&54.97&0.07\%&29s&382.36&0.58\%&31s&1629.09&1.25\%&1m30s\\
SAWT$\{4\text{k}\}$&11.84&0.42\%&31s&54.80&-0.23\%&58s&381.74&0.43\%&1m5s&1622.39&0.84\%&2m54s \\
SAWT$\{10\text{k}\}$ &\textbf{11.79}&\textbf{0.00\%}&2m30s&\textbf{54.63}&\textbf{-0.54\%}&2m32s&\textbf{379.96}&\textbf{-0.04\%}&2m40s&\textbf{1617.23}&\textbf{0.52}\%&7m32s\\
\hline
\end{tabular}
\end{sc}
\label{Main results}
\end{table*}

\subsection{The Decoder}


The decoder processes the SAWT encoder's output, denoted as $\{h_i^{(L)}\}_{i=1}^n$ and $\{h_i^{\ast (L)}\}_{i=1}^n$, as input and produces the probability of action pairs $\pi_\theta(A|\Bar{\sigma})$ and the value of the current state $V_\phi(\Bar{\sigma})$. Each of these components will be explained in detail.

\textbf{Policy decoder.} Our policy decoder primarily comprises a \textbf{Max-Pooling} sub-layer and an Action Selection Network (\textbf{AS-Net}) sub-layer. In alignment with \cite{DBLP:conf/iclr/BelloPL0B17}, our \textbf{AS-Net} employs the chain rule to factorize the probability of swap operations:
\begin{equation}
    \pi_\theta (A=(a_1, a_2) | \Bar{\sigma}) = p_{\theta_2} (a_2|a_1, \Bar{\sigma}) p_{\theta_1} (a_1| \Bar{\sigma}),
\end{equation}
where $\theta = (\theta_1, \theta_2)$ are learnable parameters and $p_{\theta_1}, p_{\theta_2}$ are MLPs followed by a softmax functions elaborated in Equation (\ref{AS-Net}), namely
\begin{equation}
\label{AS-Net} 
\begin{gathered}
    p_{\theta_1}(a_1 | \Bar{\sigma}) = \text{MLP}_{\theta_1}(h_v^{\ast} || \{h_i^{(L)}\}_{i=1}^n), \\
    p_{\theta_2} (a_2|a_1, \Bar{\sigma}) = \text{MLP}_{\theta_2}(h_v^{\ast} ||\{h_i^{(L)}\}_{i=1}^n || h_{a1}^{(L)}),
\end{gathered}
\end{equation}
where $h_v^{\ast} = \textbf{Max}(\{h_i^{\ast (L)}\}_{i=1}^n)$ and $\text{MLP}_{\theta_1}$, $\text{MLP}_{\theta_2}$ consist of three hidden layers each followed by a ReLU activation function. $\textbf{Max}$ is the $\textbf{Max-Pooling}$ operation that extracts the global representation of the best-found solution $\sigma^{\ast}$.

\textbf{Value decoder.} Our value decoder aims to evaluate the current state $\Bar{\sigma} = (\sigma,\sigma^{\ast})$,
\begin{equation}
    V_{\phi}(\Bar{\sigma}) = \mathbf{W}_2\text{ReLU}(\mathbf{W}_1(\frac{1}{n} \sum_{i=1}^n h_i^{(L)} + h_v^{\ast}) + \mathbf{b}_1) + \mathbf{b}_2,
\end{equation}
where $\mathbf{W}_1 \in \mathbb{R}^{d \times d}, \mathbf{W}_2 \in \mathbb{R}^{1 \times d}, \mathbf{b}_1 \in \mathbb{R}^d, \mathbf{b}_2 \in \mathbb{R}^1$ are all learnable parameters. We use mean-pooling to represent the global embedding of the solution $\sigma$. 

\subsection{Reinforcement Learning Training Algorithm}
\label{RL}


We maximize the expected rewards given a state $\Bar{\sigma}$, defined as $J(\theta|\Bar{\sigma}) = \mathbb{E}{\pi_\theta}[G_t | \Bar{\sigma}]$, through policy gradient optimization, following the approach in~\cite{d2020learning}. During training, with a distribution of the state $\Sigma$, we optimize the objective function $J(\theta) = \mathbb{E}_{\Bar{\sigma} \sim \Sigma}[J(\theta|\Bar{\sigma})]$. The training scheme closely resembles REINFORCE \cite{DBLP:journals/ml/Williams92}, and we optimize our policy using the approximate gradient below:
\begin{equation}
\label{update rl}
    \nabla_\theta J(\theta) = \frac{1}{B} \frac{1}{T}[\sum_{b=1}^B \sum_{t=0}^{T-1} \nabla_\theta \text{log}(\pi_\theta(A_t^b | \Bar{\sigma}_t^b)) (G_t^b - V_\phi(\Bar{\sigma}_t^b))],
\end{equation}
where $G^b_t = \sum_{t^{'} = t}^{T+t-1} \gamma^{t^{'} - t}r^b(t^{'})$. $T$ is the bootstrapping step within an episode. 

To augment the exploration capability of our agent, we include an entropy term the same as \cite{DBLP:journals/corr/SchulmanWDRK17} and \cite{d2020learning}:
\begin{equation}
    H(\theta) = \frac{1}{B} \sum_{b=1}^B \sum_{t=0}^T H(\pi_\theta(\cdot | \Bar{\sigma}_t^b)),
\end{equation}
where $H(\pi_\theta(\cdot | \Bar{\sigma}_t^b)) = -\mathbb{E}_{\pi_\theta}[\text{log} \pi_\theta(\cdot | \Bar{\sigma}_t^b)]$. Finally, we optimize the value function using the loss:
\begin{equation}
    \mathcal{L}(\phi) = \frac{1}{B} \sum_{b=1}^B \sum_{t=0}^{T-1} ||G_t^b - V_\phi(\Bar{\sigma}_t^b)||_2^2.
\end{equation}

More architecture details can be found in Appendix \ref{apend_net}.

\section{Experiments}

This section details the evaluation setup and presents the results. Additional experiments are provided in Appendix~\ref{apend_exp}. The corresponding code and other resources are released at \href{https://github.com/PKUTAN/SAWT}{https://github.com/PKUTAN/SAWT}.

\subsection{Experimental Setup}
\label{instance generation}

\paragraph{Instance generation.} We evaluate our method across four benchmark tasks: Koopmans-Beckmann's QAP with 10, 20, 50, and 100 nodes, denoted as QAP10, QAP20, QAP50, and QAP100, respectively. In all tasks, location node coordinates are uniformly randomly drawn from the unit square $[0,1]^2$. The flow $f_{ij}$ from facility $i$ to facility $j$ is sampled uniformly at random from $[0,1]$ with $f_{ij} = f_{ji}$. We then set the diagonal elements to zero and randomly set $f_{ij} = f_{ji} = 0$ with a probability $p$. For all experiments, we use a training set of up to 5120 instances and evaluate results on a test set of 256 different instances from the same distribution. We set $p = 0.7$ for all tasks.


\textbf{QAPLIB Benchmark.} The QAPLIB comprises 134 real-world QAP instances spanning 15 categories, such as planning hospital facility layouts. Table~\ref{QAPLIB} displays each category name, like ``bur," along with the instance size indicated within brackets. Each instance comprises a flow matrix and a distance matrix, with different distributions of flow and distance metrics across categories. To ensure a fair comparison, we adhere to the same settings as~\cite{liu2022revocable}. Additional statistical details of QAPLIB can be found in Appendix~\ref{apend_QAPLIB}.

\textbf{Baselines.} For the self-generated  Koopmans-Beckmann's QAP tasks, we compare our proposed method SAWT with four categories of baselines: (1) \textbf{Exact solvers}:  Gurobi \cite{gurobi2021gurobi} and Tabu-search (TS)\cite{zhang2020hybrid} (2) \textbf{Heuristic solvers}: SM \cite{leordeanu2005spectral}, RRWM \cite{cho2010reweighted}, and IPFP \cite{leordeanu2009integer}. (3) \textbf{L2I methods}: MatNet$^{\ast}$ \cite{DBLP:conf/nips/KwonCYPPG21}, Costa \cite{d2020learning}, Sui \cite{sui2021learning}, Wu \cite{wu2021learning}, Dact \cite{ma2021learning}, and NeuOpt \cite{ma2023learning}. (4) \textbf{L2C methods}: RGM \cite{liu2022revocable}. For QAPLIB benchmarks, following \cite{liu2022revocable}, the baslines are (1) \textbf{Heuristic solvers}: SM, RRWM, and SK--JA \cite{kushinsky2019sinkhorn} and (2) \textbf{Neural solvers}: NGM \cite{wang2021neural} and RGM \cite{liu2022revocable}. The baseline details are in Appendix \ref{apend_base}.

\begin{table*}[t]
\centering
\caption{Generalization experiments using a policy pre-trained on QAP50, applied to instances from QAPLIB. The mean/max/min
gaps are reported for each category. The mean performance over all categories and the inference time (s)
per instance are reported.}
\scriptsize
\begin{sc}
\begin{tabular}{l|ccc|ccc|ccc|ccc}
\hline & \multicolumn{3}{|c|}{ bur (26) } & \multicolumn{3}{|c|}{chr(12-25)} & \multicolumn{3}{|c|}{ esc (16-64)} & \multicolumn{3}{|c}{ had (12 - 20) }  \\
\cline{2-13} & Mean$\downarrow$ & Min$\downarrow$  &  Max$\downarrow$  & Mean$\downarrow$ & Min$\downarrow$  &  Max$\downarrow$ & Mean$\downarrow$ & Min$\downarrow$  &  Max$\downarrow$  & Mean$\downarrow$ & Min$\downarrow$  &  Max$\downarrow$ \\
\hline SM & 22.3 & 20.3 & 24.9 & 460.1 & 144.6 & 869.1 & 301.6 & 0.0 & 3300.0 & 17.4 & 14.7 & 21.5 \\
 RRWM & 23.1 & 19.3 & 27.3 & 616.0 & 120.5 & 1346.3 & 63.9 & 0.0 & 200.0 & 25.1 & 22.1 & 28.3 \\
 SK-JA & 4.7 & 2.8 & 6.2 & \textbf{38.5} & \textbf{0.0} &\textbf{186.1} & 364.8 & 0.0 & 2200.0 & 25.8 & 6.9 & 100.0  \\
 NGM & 3.4 & 2.8 & 4.4 & 121.3 & 45.4 & 251.9 & 126.7 & 0.0 & 200.0 & 8.2 & 6.0 & 11.6  \\
  RGM & 7.1 & 4.5 & 9.0 & 112.4 & 23.4 & 361.4 & 32.8 & 0.0 & 141.5 & 6.2 & 1.9 & 9.0 \\
\hline 
SAWT &\textbf{2.8}&\textbf{2.2}&\textbf{3.4}&110.7&8.6&201.2&\textbf{13.5}&\textbf{0.0}&\textbf{63.7}&\textbf{3.8}&\textbf{1.6}&\textbf{6.5} \\
\hline
\hline
&\multicolumn{3}{|c|}{ kra (30-32)} & \multicolumn{3}{|c|}{ lipa (20-60)} & \multicolumn{3}{|c|}{ nug (12-30)} & \multicolumn{3}{|c}{ rou (12-30)} \\
\cline{2-13}
& Mean$\downarrow$ & Min$\downarrow$  &  Max$\downarrow$  & Mean$\downarrow$ & Min$\downarrow$  &  Max$\downarrow$ & Mean$\downarrow$ & Min$\downarrow$  &  Max$\downarrow$  & Mean$\downarrow$ & Min$\downarrow$  &  Max$\downarrow$\\
\hline SM & 65.3 & 63.8 & 67.3 & 19.0 & 3.8 & 34.8 & 45.5 & 34.2 & 64.0 & 35.8 & 30.9 & 38.2 \\
RRWM  &58.8 & 53.9 & 67.7 & 20.9 & 3.6 & 41.2 & 67.8 & 52.6 & 79.6 & 51.2 & 39.3 & 60.1 \\
SK-JA & 41.4 & 38.9 & 44.4 & \textbf{0.0} & 0.0 & \textbf{0.0} & 25.3 & 10.9 & 100.0 & 13.7 & 10.3 & 17.4 \\
 NGM & 31.6 & 28.7 & 36.8 & 16.2 & 3.6 & 29.4 & 21.0 & 14.0 & 28.5 & 30.9 & 23.7 & 36.3 \\
 RGM & \textbf{15.0} & \textbf{10.4} & \textbf{20.6} & 13.3 & 3.0 & 23.8 & 9.7 & 6.1 & \textbf{12.9} & 13.4 & \textbf{7.1} & 16.7 \\
 \hline
SAWT &30.1&28.1&34.2& 0.4&\textbf{0.0}&2.6&\textbf{9.2}&\textbf{4.1}&13.1&\textbf{10.8}&8.2&\textbf{13.1} \\
\hline
\hline
&\multicolumn{3}{|c|}{ scr (12-20)} & \multicolumn{3}{|c|}{ sko(42-64)} & \multicolumn{3}{|c|}{ ste(36)} & \multicolumn{3}{|c}{ tai(12-64)} \\
\cline{2-13}
& Mean$\downarrow$ & Min$\downarrow$  &  Max$\downarrow$  & Mean$\downarrow$ & Min$\downarrow$  &  Max$\downarrow$ & Mean$\downarrow$ & Min$\downarrow$  &  Max$\downarrow$  & Mean$\downarrow$ & Min$\downarrow$  &  Max$\downarrow$\\
\hline SM & 123.4 & 104.0 & 139.1 & 29.0 & 26.6 & 31.4 & 475.5 & 197.7 & 1013.6 & 180.5 & 21.6 & 1257.9 \\
RRWM  & 173.5 & 98.9 & 218.6 & 48.5 & 47.7 & 49.3 & 539.4 & 249.5 & 1117.8 & 197.2 & 26.8 & 1256.7 \\
SK-JA & 48.6 & 44.3 & 55.7 & 18.3 & 16.1 & 20.5 & 120.4 & 72.5 & 200.4 & 25.2 & 1.6 & 107.1\\
 NGM & 55.5 & 41.4 & 66.2 & 25.2 & 22.8 & 27.7 & 101.7 & 57.6 & \textbf{172.8} & 61.4 & 18.7 & 352.1 \\
 RGM & 45.5 & 30.2 & 56.1 & \textbf{10.6} & \textbf{9.9} & \textbf{11.2} & 134.1 & 69.9 & 237.0 & 17.3 & 11.4 & \textbf{28.6} \\
 \hline
SAWT &\textbf{28.5}&\textbf{10.2}&\textbf{48.0}&17.7&16.7&18.7&\textbf{93.5}&\textbf{46.7}&\textbf{170.2}&\textbf{16.5}&\textbf{0.5}&48.7 \\
\hline
\hline
&\multicolumn{3}{|c|}{ tho (30-40)} & \multicolumn{3}{|c|}{ wil(50)} & \multicolumn{3}{|c|}{ \textbf{Average(12-64)}} & \multicolumn{3}{|c}{Time per instance} \\
\cline{2-10}
& Mean$\downarrow$ & Min$\downarrow$  &  Max$\downarrow$  & Mean$\downarrow$ & Min$\downarrow$  &  Max$\downarrow$ & Mean$\downarrow$ & Min$\downarrow$  &  Max$\downarrow$   &  \multicolumn{3}{|c}{(in seconds)} \\
\hline SM & 55.0 & 54.0 & 56.0 & 13.8 & 11.7 & 15.9 & 181.2 & 46.9 & 949.9 & & \textbf{0.01}& \\
RRWM & 80.6 & 78.2 & 83.0 & 18.2 & 12.5 & 23.8 & 169.5 & 49.5 & 432.9 & & 0.15& \\
SK-JA & 32.9 & 30.6 & 35.3 & 8.8 & 6.7 & 10.7 & 93.2 & \textbf{9.0} & 497.9 & & 563.4& \\
 NGM & 27.5 & 24.8 & 30.2 & 10.8 & 8.2 & 11.1 & 62.4 & 17.8 & 129.7 & & 15.72& \\
 RGM & \textbf{20.7} & \textbf{12.7} & 28.6 & 8.1 & 7.9 & \textbf{8.4} & 35.8 & 10.7 & 101.1 & & 75.53& \\
 \hline
SAWT &24.8&23.2&\textbf{26.4}&\textbf{8.1}&\textbf{7.6}&8.6&\textbf{26.8}&11.2&\textbf{47.0}&&12.11& \\
\hline
\end{tabular}   
\end{sc}
\label{QAPLIB}
\end{table*}


For heuristic solvers, SM, RRWM, and IPFP are renowned for matching problems. For the L2I methods, given the absence of prior work on QAP, we derive baselines from the MILP community, with a particular emphasis on VRPs. As for L2C solvers, RGM is the state-of-the-art RL method for solving QAPLIB. To demonstrate the effectiveness of our model, we slightly adapt the L2I methods to suit QAP by replacing their embeddings with those from our model. It is noteworthy that MatNet, initially an L2C method, is modified into an L2I method, denoted as MatNet$^{\ast}$. Detailed implementation information can be found in Appendix~\ref{apend_mat}.


In contrast to MILP, efficiently solving the QAP to optimality is a challenging task. Despite the rapid optimization capabilities of commercial solvers like Gurobi, SCIP~\cite{bestuzheva2021scip}, and LKH~\cite{helsgaun2017lkh} for MILP, the QAP presents a different scenario. Currently, there is no commercial solver that can efficiently handle the QAP to optimal, even for instances as small as a size of 20. Therefore, as a compromise between performance and efficiency, we have implemented the Tabu-search algorithm~\cite{zhang2020hybrid}. We conduct a search step of 5000 to acquire the Best-Known Solution (\textbf{BKS}) for comparative analysis. Specifically, for QAP10, we leverage Gurobi to solve instances optimally. For QAP20 and QAP50, we limit the solution time to 10 minutes per instance. However, when addressing QAP100, Gurobi reported a segmentation fault on our hardware, primarily attributed to constraints in CPU memory capacity.


\textbf{Metrics.} We assess all approaches using two key metrics: (1) The \textbf{Mean} objective value, calculated as the average over all test instances using Equation~\ref{trace QAP}. (2) The \textbf{Gap}, representing the difference between the \textbf{Mean} objective value of the approaches and the \textbf{Mean}$^{\ast}$ objective value of the BKS, defined as $Gap = \frac{Mean - Mean^{\ast}}{Mean^{\ast}}$.


\textbf{Implementation.} Our model SAWT is implemented using PyTorch~\cite{paszke2019pytorch}. All experiments are executed on a single NVIDIA 3080Ti GPU (12GB) and a 12th Gen Intel(R) Core(TM) i5-12600KF 3.69 GHz CPU. We train the model for 200 epochs with a batch size of 512 and an episode length of 400 for all tasks. Each epoch requires an average time of 1m40s, 4m20s, 9m54s, and 19m48s for QAP10, QAP20, QAP50, and QAP100, respectively. During testing, our policy runs for 2000, 4000, and 10000 steps on 256 instances, and we use the averaged objective value for comparison with other baselines.

\subsection{Experimental Results}


\textbf{Results on Koopmans-Beckmann's QAP.} Table \ref{Main results} presents the results for Koopmans-Beckmann's QAP. It is evident that QAP10 is relatively straightforward, with Gurobi achieving optimal solutions for all 256 testing instances within 5 minutes. However, Gurobi faces significant challenges with QAP100, reaching the limits of CPU memory capacity. These Gurobi-based outcomes underscore the inherent complexities associated with solving the QAP. Additionally, the performance issues encountered by all three heuristics further emphasize the difficulty of QAP.

Regarding the L2I methods, there are three main observations: (1) All L2I methods exhibit poor performance. The best gaps on four tasks are 2.1\%, 3.32\%, 4.24\%, and 5.01\%  which are almost an average of 10 $\times$  larger than the gaps achieved by our SAWT method at the 2000-step search. This is because they are designed to capture the linear structure of the VRPs that can not learn the higher-order patterns in the QAPs. (2) MatNet$^{\ast}$ displays notably poor performance, significantly trailing our SAWT by margins of $0.67\%$ and $0.07\%$ even with a search step of 2000. This suggests the effectiveness of our SAWT to encode the facility and location nodes independently. Furthermore, the dual attention structure of MatNet$^{\ast}$ leads to high computational complexity, resulting in its inability to solve larger instances like QAP50 and QAP100. (3) Our SAWT model demonstrates superior performance over all heuristic and L2I baselines, even with a search step constraint of 2000, showing a strong ability to learn the patterns of the QAP. This leads to even smaller gaps of $0.00\%$ (QAP10), $-0.45\%$ (QAP20), $-0.04\%$ (QAP50), and $0.52\%$ (QAP100) when the search step is increased to 10,000.


In L2C methods, RGM outperforms all L2I methods on QAP10 and QAP20, achieving gaps of $1.78\%$ and $1.87\%$, respectively. This success stems from RGM's learning strategy, enabling it to grasp more complex patterns within the QAP. However, severe scalability issues plague RGM, making it incapable of solving QAP instances beyond a size of 50. This substantial limitation, arising from constraints associated with the association graph, severely restricts RGM's potential for real-world applications.


In terms of inference time, our SAWT model demonstrates comparable efficiency to Dact and slightly outperforms other L2I methods. Notably, with a search step of 10,000, our model achieves significantly faster inference times than the exact solver Tabu-search with a step size of 5,000. Specifically, it is $4.7 \times$ faster for QAP10, $38.7 \times$ for QAP20, $403 \times$ for QAP50, and an impressive $5350\times$ faster for QAP100. In summary, our SAWT model excels in both effectiveness and efficiency when solving the QAP.

\begin{table}[t]
\centering
\caption{Generalization to different sizes of the QAP instances.}
\begin{sc}
 \begin{tabular}{c|ccc}
\hline
Methods & QAP20 & QAP50 & QAP100\\
\hline
Policy-20 & \textbf{54.63} & 383.67& 1641.40 \\

Policy-50 & 54.78& \textbf{380.38} & 1617.84\\
Policy-100 & 54.82 & 381.13 & \textbf{1617.23}\\
\hline
\end{tabular}
\label{Generalize:1}   
\end{sc}
\vspace{-0.15in}
\end{table}


\textbf{Generalization to Different Sizes.} We evaluate the performance of our policies trained on QAP20, QAP50, and QAP100 when applied to QAP tasks of varying sizes. Since our facility nodes
embedding design, as outlined in section \ref{embeddings}, adopt a strategy of randomly choosing a $N_{init}$-dimensional one-hot vector for the initial embeddings, we can develop generalized
policies capable of solving the QAP for any instance with a
size less than $N_{init}$, we set $N_{init} = 128$ for our experiments. As depicted in Table~\ref{Generalize:1}, all our policies demonstrate the ability to generalize effectively across different task sizes. Among all, Policy-50 exhibits superior generalization compared to other policies. Surpassing Policy-100 in QAP20 performance with a score of 54.78 versus 54.82, Policy-50 maintains competitive effectiveness on QAP100, achieving a score of 1617.84 against 1617.23, reflecting a marginal difference of only $0.03\%$.


\textbf{Generalization to QAPLIB.} We extend our experiments to evaluate the SAWT model on the QAPLIB benchmark, utilizing Policy-50 for solving instances from QAPLIB. The results, summarized in Table \ref{QAPLIB}, show our model's superiority, surpassing all baseline metrics with an average mean of $26.8\%$ and a maximum value of $47.0\%$. Additionally, our model outperforms learning methods such as RGM and NGM in terms of inference speed. Noteworthy observations from Table \ref{QAPLIB} reveal that while the SAWT method excels in performance on datasets like ``Bur" and ``Lipa," it faces challenges on datasets such as ``Chr" and ``Ste," evidenced by substantial Mean gaps of $110.7\%$ and $93.5\%$. This performance variation can be attributed to specific dataset characteristics, including the extreme sparsity of the flow matrix in ``Chr" and the distance matrix in ``Ste," resulting in a significant distribution shift compared to our training dataset. Further details on QAPLIB experiments are available in Appendix~\ref{qaplib_append}.

\begin{table}[t]
\centering
\caption{Ablation studies on 256 QAP20 instances running policies for 10000 steps.}
\begin{sc}
 \begin{tabular}{cc|cc}
\hline
Single-AM& SAWT-encoder & Mean & Gap \\
\hline
$\times$ & $\times$ & 62.38 & 13.56\%\\
$\times$ & $\surd$ & 61.89 & 12.67\%\\
$\surd$ & $\times$ & 57.00 & 3.76\%\\
$\surd$ & $\surd$ & 54.63  & -0.54\%\\
\hline
\end{tabular}
\label{Ablation}   
\end{sc}
\vspace{-0.1in}
\end{table}

\begin{table*}[t]
\centering
\caption{Random initial solutions experiments on QAP tasks.}
\small
\begin{sc}
\begin{tabular}{l|cc|cc|cc}
\hline \multirow{2}{*}{Methods}  & \multicolumn{2}{|c|}{QAP20} & \multicolumn{2}{|c|}{QAP50}&  \multicolumn{2}{|c}{QAP100} \\
\cline{2-7} & Mean &Gap & Mean &Gap & Mean &Gap   \\
\hline
SAWT&54.63 $\pm$ 0.00&-0.54\% $\pm$ 0.01&379.97 $\pm$ 0.12& -0.04\% $\pm$ 0.03&1617.08 $\pm$ 2.84 &0.51\% $\pm$ 0.12\\
SAWT(rand) &54.69  $\pm$ 0.02 & -0.43\% $\pm$ 0.03 &380.83 $\pm$ 0.72 & 0.18\% $\pm$ 0.18&1622.06 $\pm$ 4.21 & 0.82\% $\pm$  0.26\\

SAWT-rand &\textbf{54.62}  $\pm$ 0.00 & \textbf{-0.55}\% $\pm$ 0.03 &\textbf{379.35} $\pm$ 0.06  & \textbf{-0.38}\% $\pm$ 0.01& \textbf{1614.42} $\pm$ 1.03 & \textbf{0.34}\% $\pm$  0.07\\
\hline
\end{tabular}
\end{sc}
\label{random init}
\end{table*}

\begin{table*}[t]
\centering
\scriptsize
\caption{Random initial solutions experiments on QAPLIB.}
\begin{sc}
\begin{tabular}{l|c|c|c|c|c|c|c}
\hline
   & bur&chr&esc&had&kra&lipa&nug \\
\hline
SAWT & 2.8\% $\pm$ \textbf{0.0} &108.6\% $\pm$ 8.2 &14.2\% $\pm$ 1.4&3.7\% $\pm$ \textbf{0.2}&31.2\% $\pm$ 1.7&0.4\% $\pm$ \textbf{0.0}&9.1\%  $\pm$ \textbf{0.2} \\
SAWT(rand) & 3.3\% $\pm$ 0.3 &128.2\% $\pm$ 29.4 &13.1\% $\pm$ 2.1&4.5\% $\pm$ 0.5&42.2\% $\pm$ 14.2&0.7\% $\pm$ 0.2&11.3\%  $\pm$ 1.9 \\
SAWT-rand & \textbf{2.6}\% $\pm$ 0.1 &\textbf{84.6}\% $\pm$ \textbf{3.1} &\textbf{12.9}\% $\pm$ \textbf{0.6}& \textbf{3.4}\% $\pm$ 0.3&\textbf{28.7}\% $\pm$ \textbf{1.4}&\textbf{0.4}\% $\pm$ 0.1&\textbf{9.0}\%  $\pm$ 0.3 \\
\hline
&rou&scr&sko&ste&tai&tho&wil \\
\hline
SAWT & 10.5\% $\pm$ \textbf{0.5}&\textbf{29.0}\% $\pm$ \textbf{1.1}&17.7\%  $\pm$ 0.9&98.2\% $\pm$ \textbf{5.1}&16.3\% $\pm$ \textbf{0.3} &24.3\% $\pm$ \textbf{1.1} & 8.1 $\pm$ \textbf{0.5}\\
SAWT(rand) & 9.2\% $\pm$ 1.7&32.8\% $\pm$ 4.4&15.3\%  $\pm$ 5.3&124.5\% $\pm$ 20.7&16.4\% $\pm$ 0.8 &24.1\% $\pm$ 2.3 & 10.3 $\pm$ 2.6\\
SAWT-rand & \textbf{8.5}\% $\pm$ 0.9 &29.5\% $\pm$ 2.4 &\textbf{13.4}\% $\pm$ \textbf{0.7}&\textbf{75.6}\% $\pm$ 9.5&\textbf{16.0}\% $\pm$ 0.4&\textbf{20.2}\% $\pm$ 1.2&\textbf{7.9}\%  $\pm$ 0.9 \\
\hline
\end{tabular}
\end{sc}
\label{QAPLIB random}
\end{table*}

\subsection{Additional Analysis}

\textbf{Ablation studies.} In Table \ref{Ablation}, we present an ablation study of the proposed method: SAWT. ``Single-AM" implies that we only use a single self-attention model and a GCN to encode the facility and location nodes, which is aligned with our formal design, otherwise, we use a dual self-attention method as in MatNet. As shown in the table, both ``Single-AM" and ``SAWT-encoder" play an important role in our model. Without ``Single-AM", the model struggles to tackle the QAP, this proves the fact that we need to process facility and location nodes differently without early interaction as they are from different domains. ``SAWT-encoder" is also indispensable, by equipping our model with ``SAWT-encoder", the gaps lowered from $13.56\%$ to $11.94\%$ and from $3.78\%$ to $-0.45\%$. In conclusion, we consider``Single-AM" to be more crucial than ``SAWT-encoder" in our model. This is because without meaningful representations, the model faces significant challenges in effectively solving the QAP.



\textbf{Robustness against initial solution.} In Tables \ref{random init} and \ref{QAPLIB random}, we assess our model's robustness against random initial solutions for QAP tasks and QAPLIB. ``SAWT (RAND)" refers to SAWT trained with a fixed initial solution but evaluated with random ones, while ``SAWT-RAND" is both trained and evaluated with random initial solutions. Two key observations emerge: (1) SAWT (RAND) performs the worst, and SAWT-RAND the best, for both QAP tasks and QAPLIB. This can be attributed to the ``explore and exploit" principle of RL. By using diverse initial solutions, SAWT-RAND effectively explores the solution space, thereby improving its performance. (2) The variance in Table \ref{QAPLIB random} for SAWT-RAND is generally lower than for SAWT (RAND) but higher than for SAWT. This results from their initial solution strategies: SAWT uses fixed initial solutions, while both SAWT-RAND and SAWT (RAND) use random ones. In summary, SAWT-RAND demonstrates superior performance on both QAP tasks and QAPLIB, highlighting our model's robustness against random initial solutions.


\section{Conclusions}

In this paper, we introduce SAWT, a novel L2I solver for Koopmans-Beckmann's QAP, the first L2I method to solve the complex QAP. It first processes the facility and location nodes independently by a self-attention model and a GCN. It is then followed by an SAWT encoder to capture the rich patterns in QAP. Extensive experiments on both synthetic and benchmark datasets justified the effectiveness of SAWT in terms of both inference and generalization. Additional analysis furthermore demonstrates the robustness of SAWT. It is future work to learn a model that has stronger generalization ability. Although SAWT achieves the best performance on the QAPLIB, it still has a relatively high mean gap. This is mainly because of the diverse distribution of instances in the QAPLIB. In the future, it is desirable to utilize a meta-learning strategy to enhance SAWT, so that it can perform better on QAPLIB.

\section*{Acknowledgement} This research work is supported by National Key R\&D Program of China (2022ZD0160305), a research grant from China Tower Corporation Limited, and a grant from Beijing Aerospace Automatic Control Institute.

\section*{Impact Statement}

This work boosts the scalability, performance, and efficiency of solving QAPs but necessitates vigilant data safety measures due to its wide real-world applicability in areas such as chip placement and hospital layouts. The risk of data leakage looms, particularly with sensitive datasets like chip circuits or confidential hospital patient data. Currently, as we focus on learning-based methods with self-generated tasks, data safety concerns are not immediate. However, the future potential for data safety issues warrants attention.

\nocite{langley00}

\bibliography{example_paper}

\begin{thebibliography}{59}
\providecommand{\natexlab}[1]{#1}
\providecommand{\url}[1]{\texttt{#1}}
\expandafter\ifx\csname urlstyle\endcsname\relax
  \providecommand{\doi}[1]{doi: #1}\else
  \providecommand{\doi}{doi: \begingroup \urlstyle{rm}\Url}\fi

\bibitem[Ahmed(2015)]{ahmed2015multi}
Ahmed, Z.~H.
\newblock A multi-parent genetic algorithm for the quadratic assignment problem.
\newblock \emph{Opsearch}, 52:\penalty0 714--732, 2015.

\bibitem[Ahn et~al.(2020)Ahn, Kim, Lee, and Shin]{DBLP:conf/nips/AhnKLS20}
Ahn, S., Kim, J., Lee, H., and Shin, J.
\newblock Guiding deep molecular optimization with genetic exploration.
\newblock In Larochelle, H., Ranzato, M., Hadsell, R., Balcan, M., and Lin, H. (eds.), \emph{Advances in Neural Information Processing Systems 33: Annual Conference on Neural Information Processing Systems 2020, NeurIPS 2020, December 6-12, 2020, virtual}, 2020.

\bibitem[Ba et~al.(2016)Ba, Kiros, and Hinton]{DBLP:journals/corr/BaKH16}
Ba, L.~J., Kiros, J.~R., and Hinton, G.~E.
\newblock Layer normalization.
\newblock \emph{CoRR}, abs/1607.06450, 2016.

\bibitem[Bello et~al.(2017)Bello, Pham, Le, Norouzi, and Bengio]{DBLP:conf/iclr/BelloPL0B17}
Bello, I., Pham, H., Le, Q.~V., Norouzi, M., and Bengio, S.
\newblock Neural combinatorial optimization with reinforcement learning.
\newblock In \emph{5th International Conference on Learning Representations, {ICLR} 2017, Toulon, France, April 24-26, 2017, Workshop Track Proceedings}. OpenReview.net, 2017.

\bibitem[Bestuzheva et~al.(2021)Bestuzheva, Besan{\c{c}}on, Chen, Chmiela, Donkiewicz, van Doornmalen, Eifler, Gaul, Gamrath, Gleixner, et~al.]{bestuzheva2021scip}
Bestuzheva, K., Besan{\c{c}}on, M., Chen, W.-K., Chmiela, A., Donkiewicz, T., van Doornmalen, J., Eifler, L., Gaul, O., Gamrath, G., Gleixner, A., et~al.
\newblock The scip optimization suite 8.0.
\newblock \emph{arXiv preprint arXiv:2112.08872}, 2021.

\bibitem[Burkard et~al.(1997)Burkard, Karisch, and Rendl]{burkard1997qaplib}
Burkard, R.~E., Karisch, S.~E., and Rendl, F.
\newblock Qaplib--a quadratic assignment problem library.
\newblock \emph{Journal of Global optimization}, 10:\penalty0 391--403, 1997.

\bibitem[Chen \& Tian(2019)Chen and Tian]{chen2019learning}
Chen, X. and Tian, Y.
\newblock Learning to perform local rewriting for combinatorial optimization.
\newblock \emph{Advances in Neural Information Processing Systems}, 32, 2019.

\bibitem[Cho et~al.(2010)Cho, Lee, and Lee]{cho2010reweighted}
Cho, M., Lee, J., and Lee, K.~M.
\newblock Reweighted random walks for graph matching.
\newblock In \emph{Computer Vision--ECCV 2010: 11th European Conference on Computer Vision, Heraklion, Crete, Greece, September 5-11, 2010, Proceedings, Part V 11}, pp.\  492--505. Springer, 2010.

\bibitem[Costa et~al.(2020)Costa, Rhuggenaath, Zhang, and Akcay]{d2020learning}
Costa, P.~R., Rhuggenaath, J., Zhang, Y., and Akcay, A.
\newblock Learning 2-opt heuristics for the traveling salesman problem via deep reinforcement learning.
\newblock In \emph{Asian conference on machine learning}, pp.\  465--480. PMLR, 2020.

\bibitem[Cui et~al.(2023)Cui, Baumg{\"a}rtner, Yilmaz, Li, Fabian, Becker, Xiang, Bauer, and Koeppl]{cui2023uav}
Cui, K., Baumg{\"a}rtner, L., Yilmaz, M.~B., Li, M., Fabian, C., Becker, B., Xiang, L., Bauer, M., and Koeppl, H.
\newblock Uav swarms for joint data ferrying and dynamic cell coverage via optimal transport descent and quadratic assignment.
\newblock In \emph{2023 IEEE 48th Conference on Local Computer Networks (LCN)}, pp.\  1--8. IEEE, 2023.

\bibitem[Dai et~al.(2016)Dai, Dai, and Song]{dai2016discriminative}
Dai, H., Dai, B., and Song, L.
\newblock Discriminative embeddings of latent variable models for structured data.
\newblock In \emph{International conference on machine learning}, pp.\  2702--2711. PMLR, 2016.

\bibitem[Edwards(1980)]{edwards1980branch}
Edwards, C.
\newblock A branch and bound algorithm for the koopmans-beckmann quadratic assignment problem.
\newblock \emph{Combinatorial optimization II}, pp.\  35--52, 1980.

\bibitem[Elshafei(1977)]{elshafei1977hospital}
Elshafei, A.~N.
\newblock Hospital layout as a quadratic assignment problem.
\newblock \emph{Journal of the Operational Research Society}, 28\penalty0 (1):\penalty0 167--179, 1977.

\bibitem[Gurobi(2023)]{gurobi2021gurobi}
Gurobi, L.
\newblock Gurobi optimizer reference manual, 2023.

\bibitem[Hanh et~al.(2019)Hanh, Binh, Hoai, and Palaniswami]{hanh2019efficient}
Hanh, N.~T., Binh, H. T.~T., Hoai, N.~X., and Palaniswami, M.~S.
\newblock An efficient genetic algorithm for maximizing area coverage in wireless sensor networks.
\newblock \emph{Information Sciences}, 488:\penalty0 58--75, 2019.

\bibitem[He et~al.(2016)He, Zhang, Ren, and Sun]{DBLP:conf/cvpr/HeZRS16}
He, K., Zhang, X., Ren, S., and Sun, J.
\newblock Deep residual learning for image recognition.
\newblock In \emph{2016 {IEEE} Conference on Computer Vision and Pattern Recognition, {CVPR} 2016, Las Vegas, NV, USA, June 27-30, 2016}, pp.\  770--778. {IEEE} Computer Society, 2016.
\newblock \doi{10.1109/CVPR.2016.90}.

\bibitem[Helsgaun(2017)]{helsgaun2017lkh}
Helsgaun, K.
\newblock Lkh-3 version 3.0. 6 (may 2019), 2017.

\bibitem[Hu et~al.(2022)Hu, Sun, and Yang]{hu2022switch}
Hu, Z., Sun, Y., and Yang, Y.
\newblock Switch to generalize: Domain-switch learning for cross-domain few-shot classification.
\newblock In \emph{International Conference on Learning Representations}, 2022.
\newblock URL \url{https://openreview.net/forum?id=H-iABMvzIc}.

\bibitem[Hu et~al.(2023{\natexlab{a}})Hu, Sun, Wang, and Yang]{hu2023dacdetr}
Hu, Z., Sun, Y., Wang, J., and Yang, Y.
\newblock {DAC}-{DETR}: Divide the attention layers and conquer.
\newblock In \emph{Thirty-seventh Conference on Neural Information Processing Systems}, 2023{\natexlab{a}}.
\newblock URL \url{https://openreview.net/forum?id=8JMexYVcXB}.

\bibitem[Hu et~al.(2023{\natexlab{b}})Hu, Sun, and Yang]{hu2023suppressing}
Hu, Z., Sun, Y., and Yang, Y.
\newblock Suppressing the heterogeneity: A strong feature extractor for few-shot segmentation.
\newblock In \emph{The Eleventh International Conference on Learning Representations}, 2023{\natexlab{b}}.
\newblock URL \url{https://openreview.net/forum?id=CGuvK3U09LH}.

\bibitem[James et~al.(2009)James, Rego, and Glover]{james2009cooperative}
James, T., Rego, C., and Glover, F.
\newblock A cooperative parallel tabu search algorithm for the quadratic assignment problem.
\newblock \emph{European Journal of Operational Research}, 195\penalty0 (3):\penalty0 810--826, 2009.

\bibitem[Jiang et~al.(2023)Jiang, Jin, Tan, and Mu]{jiang2023video}
Jiang, B., Jin, Y., Tan, Z., and Mu, Y.
\newblock Video action segmentation via contextually refined temporal keypoints.
\newblock In \emph{Proceedings of the IEEE/CVF International Conference on Computer Vision}, pp.\  13836--13845, 2023.

\bibitem[Khalil et~al.(2017)Khalil, Dai, Zhang, Dilkina, and Song]{khalil2017learning}
Khalil, E., Dai, H., Zhang, Y., Dilkina, B., and Song, L.
\newblock Learning combinatorial optimization algorithms over graphs.
\newblock \emph{Advances in neural information processing systems}, 30, 2017.

\bibitem[Koopmans \& Beckmann(1957)Koopmans and Beckmann]{koopmans1957assignment}
Koopmans, T.~C. and Beckmann, M.
\newblock Assignment problems and the location of economic activities.
\newblock \emph{Econometrica: journal of the Econometric Society}, pp.\  53--76, 1957.

\bibitem[Kushinsky et~al.(2019)Kushinsky, Maron, Dym, and Lipman]{kushinsky2019sinkhorn}
Kushinsky, Y., Maron, H., Dym, N., and Lipman, Y.
\newblock Sinkhorn algorithm for lifted assignment problems.
\newblock \emph{SIAM Journal on Imaging Sciences}, 12\penalty0 (2):\penalty0 716--735, 2019.

\bibitem[Kwon et~al.(2021)Kwon, Choo, Yoon, Park, Park, and Gwon]{DBLP:conf/nips/KwonCYPPG21}
Kwon, Y., Choo, J., Yoon, I., Park, M., Park, D., and Gwon, Y.
\newblock Matrix encoding networks for neural combinatorial optimization.
\newblock In Ranzato, M., Beygelzimer, A., Dauphin, Y.~N., Liang, P., and Vaughan, J.~W. (eds.), \emph{Advances in Neural Information Processing Systems 34: Annual Conference on Neural Information Processing Systems 2021, NeurIPS 2021, December 6-14, 2021, virtual}, pp.\  5138--5149, 2021.

\bibitem[Kwon et~al.(2020)Kwon, Choo, Kim, Yoon, Gwon, and Min]{kwon2020pomo}
Kwon, Y.-D., Choo, J., Kim, B., Yoon, I., Gwon, Y., and Min, S.
\newblock Pomo: Policy optimization with multiple optima for reinforcement learning.
\newblock \emph{Advances in Neural Information Processing Systems}, 33:\penalty0 21188--21198, 2020.

\bibitem[Lai et~al.(2022)Lai, Mu, and Luo]{DBLP:conf/nips/LaiM022}
Lai, Y., Mu, Y., and Luo, P.
\newblock Maskplace: Fast chip placement via reinforced visual representation learning.
\newblock In Koyejo, S., Mohamed, S., Agarwal, A., Belgrave, D., Cho, K., and Oh, A. (eds.), \emph{Advances in Neural Information Processing Systems 35: Annual Conference on Neural Information Processing Systems 2022, NeurIPS 2022, New Orleans, LA, USA, November 28 - December 9, 2022}, 2022.

\bibitem[Laporte \& Mercure(1988)Laporte and Mercure]{LAPORTE1988378}
Laporte, G. and Mercure, H.
\newblock Balancing hydraulic turbine runners: A quadratic assignment problem.
\newblock \emph{European Journal of Operational Research}, 35\penalty0 (3):\penalty0 378--381, 1988.
\newblock ISSN 0377-2217.
\newblock \doi{https://doi.org/10.1016/0377-2217(88)90227-5}.

\bibitem[Leordeanu \& Hebert(2005)Leordeanu and Hebert]{leordeanu2005spectral}
Leordeanu, M. and Hebert, M.
\newblock A spectral technique for correspondence problems using pairwise constraints.
\newblock In \emph{Tenth IEEE International Conference on Computer Vision (ICCV'05) Volume 1}, volume~2, pp.\  1482--1489. IEEE, 2005.

\bibitem[Leordeanu et~al.(2009)Leordeanu, Hebert, and Sukthankar]{leordeanu2009integer}
Leordeanu, M., Hebert, M., and Sukthankar, R.
\newblock An integer projected fixed point method for graph matching and map inference.
\newblock \emph{Advances in neural information processing systems}, 22, 2009.

\bibitem[Lin et~al.(2023)Lin, Yang, Yu, Hu, Zhang, and Peng]{lin2023graph}
Lin, Y., Yang, M., Yu, J., Hu, P., Zhang, C., and Peng, X.
\newblock Graph matching with bi-level noisy correspondence.
\newblock In \emph{Proceedings of the IEEE/CVF International Conference on Computer Vision}, pp.\  23362--23371, 2023.

\bibitem[Liu et~al.(2022)Liu, Jiang, Wang, Huang, Lu, and Yan]{liu2022revocable}
Liu, C., Jiang, Z., Wang, R., Huang, L., Lu, P., and Yan, J.
\newblock Revocable deep reinforcement learning with affinity regularization for outlier-robust graph matching.
\newblock In \emph{The Eleventh International Conference on Learning Representations}, 2022.

\bibitem[Ma et~al.(2021)Ma, Li, Cao, Song, Zhang, Chen, and Tang]{ma2021learning}
Ma, Y., Li, J., Cao, Z., Song, W., Zhang, L., Chen, Z., and Tang, J.
\newblock Learning to iteratively solve routing problems with dual-aspect collaborative transformer.
\newblock \emph{Advances in Neural Information Processing Systems}, 34:\penalty0 11096--11107, 2021.

\bibitem[Ma et~al.(2023)Ma, Cao, and Chee]{ma2023learning}
Ma, Y., Cao, Z., and Chee, Y.~M.
\newblock Learning to search feasible and infeasible regions of routing problems with flexible neural k-opt.
\newblock \emph{arXiv preprint arXiv:2310.18264}, 2023.

\bibitem[Mihi{\'c} et~al.(2018)Mihi{\'c}, Ryan, and Wood]{mihic2018randomized}
Mihi{\'c}, K., Ryan, K., and Wood, A.
\newblock Randomized decomposition solver with the quadratic assignment problem as a case study.
\newblock \emph{INFORMS Journal on Computing}, 30\penalty0 (2):\penalty0 295--308, 2018.

\bibitem[Mirhoseini et~al.(2021)Mirhoseini, Goldie, Yazgan, Jiang, Songhori, Wang, Lee, Johnson, Pathak, Nazi, Pak, Tong, Srinivasa, Hang, Tuncer, Le, Laudon, Ho, Carpenter, and Dean]{DBLP:journals/nature/MirhoseiniGYJSW21}
Mirhoseini, A., Goldie, A., Yazgan, M., Jiang, J.~W., Songhori, E.~M., Wang, S., Lee, Y., Johnson, E., Pathak, O., Nazi, A., Pak, J., Tong, A., Srinivasa, K., Hang, W., Tuncer, E., Le, Q.~V., Laudon, J., Ho, R., Carpenter, R., and Dean, J.
\newblock A graph placement methodology for fast chip design.
\newblock \emph{Nat.}, 594\penalty0 (7862):\penalty0 207--212, 2021.
\newblock \doi{10.1038/S41586-021-03544-W}.

\bibitem[Nowak et~al.(2018)Nowak, Villar, Bandeira, and Bruna]{nowak2018revised}
Nowak, A., Villar, S., Bandeira, A.~S., and Bruna, J.
\newblock Revised note on learning quadratic assignment with graph neural networks.
\newblock In \emph{2018 IEEE Data Science Workshop (DSW)}, pp.\  1--5. IEEE, 2018.

\bibitem[Paszke et~al.(2019)Paszke, Gross, Massa, Lerer, Bradbury, Chanan, Killeen, Lin, Gimelshein, Antiga, et~al.]{paszke2019pytorch}
Paszke, A., Gross, S., Massa, F., Lerer, A., Bradbury, J., Chanan, G., Killeen, T., Lin, Z., Gimelshein, N., Antiga, L., et~al.
\newblock Pytorch: An imperative style, high-performance deep learning library.
\newblock \emph{Advances in neural information processing systems}, 32, 2019.

\bibitem[Sahni \& Gonzalez(1976)Sahni and Gonzalez]{DBLP:journals/jacm/SahniG76}
Sahni, S. and Gonzalez, T.~F.
\newblock P-complete approximation problems.
\newblock \emph{J. {ACM}}, 23\penalty0 (3):\penalty0 555--565, 1976.
\newblock \doi{10.1145/321958.321975}.

\bibitem[Schulman et~al.(2017)Schulman, Wolski, Dhariwal, Radford, and Klimov]{DBLP:journals/corr/SchulmanWDRK17}
Schulman, J., Wolski, F., Dhariwal, P., Radford, A., and Klimov, O.
\newblock Proximal policy optimization algorithms.
\newblock \emph{CoRR}, abs/1707.06347, 2017.

\bibitem[Shylo(2017)]{shylo2017solving}
Shylo, P.
\newblock Solving the quadratic assignment problem by the repeated iterated tabu search method.
\newblock \emph{Cybernetics and Systems Analysis}, 53:\penalty0 308--311, 2017.

\bibitem[Steinberg(1961)]{53bbb073-5adc-3787-97ba-b0c9dae99c60}
Steinberg, L.
\newblock The backboard wiring problem: A placement algorithm.
\newblock \emph{SIAM Review}, 3\penalty0 (1):\penalty0 37--50, 1961.
\newblock ISSN 00361445.

\bibitem[Sui et~al.(2021)Sui, Ding, Liu, Xu, and Bu]{sui2021learning}
Sui, J., Ding, S., Liu, R., Xu, L., and Bu, D.
\newblock Learning 3-opt heuristics for traveling salesman problem via deep reinforcement learning.
\newblock In \emph{Asian Conference on Machine Learning}, pp.\  1301--1316. PMLR, 2021.

\bibitem[Taillard(1991)]{taillard1991robust}
Taillard, {\'E}.
\newblock Robust taboo search for the quadratic assignment problem.
\newblock \emph{Parallel computing}, 17\penalty0 (4-5):\penalty0 443--455, 1991.

\bibitem[Tan \& Mu(2024)Tan and Mu]{tan2024hierarchical}
Tan, Z. and Mu, Y.
\newblock Hierarchical reinforcement learning for chip-macro placement in integrated circuit.
\newblock \emph{Pattern Recognition Letters}, 2024.

\bibitem[Tan et~al.(2024)Tan, Wang, and Shan]{tan2024vision}
Tan, Z., Wang, W., and Shan, C.
\newblock Vision transformers are active learners for image copy detection.
\newblock \emph{Neurocomputing}, 587:\penalty0 127687, 2024.

\bibitem[Vaswani et~al.(2017)Vaswani, Shazeer, Parmar, Uszkoreit, Jones, Gomez, Kaiser, and Polosukhin]{DBLP:conf/nips/VaswaniSPUJGKP17}
Vaswani, A., Shazeer, N., Parmar, N., Uszkoreit, J., Jones, L., Gomez, A.~N., Kaiser, L., and Polosukhin, I.
\newblock Attention is all you need.
\newblock In Guyon, I., von Luxburg, U., Bengio, S., Wallach, H.~M., Fergus, R., Vishwanathan, S. V.~N., and Garnett, R. (eds.), \emph{Advances in Neural Information Processing Systems 30: Annual Conference on Neural Information Processing Systems 2017, December 4-9, 2017, Long Beach, CA, {USA}}, pp.\  5998--6008, 2017.

\bibitem[Veres \& Moussa(2020)Veres and Moussa]{DBLP:journals/tits/VeresM20}
Veres, M. and Moussa, M.
\newblock Deep learning for intelligent transportation systems: {A} survey of emerging trends.
\newblock \emph{{IEEE} Trans. Intell. Transp. Syst.}, 21\penalty0 (8):\penalty0 3152--3168, 2020.
\newblock \doi{10.1109/TITS.2019.2929020}.

\bibitem[Vinyals et~al.(2015)Vinyals, Fortunato, and Jaitly]{vinyals2015pointer}
Vinyals, O., Fortunato, M., and Jaitly, N.
\newblock Pointer networks.
\newblock \emph{Advances in neural information processing systems}, 28, 2015.

\bibitem[Wang \& Alidaee(2023)Wang and Alidaee]{wang2023new}
Wang, H. and Alidaee, B.
\newblock A new hybrid-heuristic for large-scale combinatorial optimization: A case of quadratic assignment problem.
\newblock \emph{Computers \& Industrial Engineering}, 179:\penalty0 109220, 2023.

\bibitem[Wang et~al.(2021)Wang, Yan, and Yang]{wang2021neural}
Wang, R., Yan, J., and Yang, X.
\newblock Neural graph matching network: Learning lawler’s quadratic assignment problem with extension to hypergraph and multiple-graph matching.
\newblock \emph{IEEE Transactions on Pattern Analysis and Machine Intelligence}, 44\penalty0 (9):\penalty0 5261--5279, 2021.

\bibitem[Williams(1992)]{DBLP:journals/ml/Williams92}
Williams, R.~J.
\newblock Simple statistical gradient-following algorithms for connectionist reinforcement learning.
\newblock \emph{Mach. Learn.}, 8:\penalty0 229--256, 1992.
\newblock \doi{10.1007/BF00992696}.

\bibitem[Wu et~al.(2021)Wu, Song, Cao, Zhang, and Lim]{wu2021learning}
Wu, Y., Song, W., Cao, Z., Zhang, J., and Lim, A.
\newblock Learning improvement heuristics for solving routing problems.
\newblock \emph{IEEE transactions on neural networks and learning systems}, 33\penalty0 (9):\penalty0 5057--5069, 2021.

\bibitem[Yang et~al.(2023)Yang, Zeng, Wu, and Yan]{DBLP:conf/www/YangZWY23}
Yang, N., Zeng, K., Wu, Q., and Yan, J.
\newblock Molerec: Combinatorial drug recommendation with substructure-aware molecular representation learning.
\newblock In Ding, Y., Tang, J., Sequeda, J.~F., Aroyo, L., Castillo, C., and Houben, G. (eds.), \emph{Proceedings of the {ACM} Web Conference 2023, {WWW} 2023, Austin, TX, USA, 30 April 2023 - 4 May 2023}, pp.\  4075--4085. {ACM}, 2023.
\newblock \doi{10.1145/3543507.3583872}.

\bibitem[Yu et~al.(2021)Yu, Wang, Yan, and Li]{yu2021deep}
Yu, T., Wang, R., Yan, J., and Li, B.
\newblock Deep latent graph matching.
\newblock In \emph{International Conference on Machine Learning}, pp.\  12187--12197. PMLR, 2021.

\bibitem[Zanfir \& Sminchisescu(2018)Zanfir and Sminchisescu]{DBLP:conf/cvpr/ZanfirS18}
Zanfir, A. and Sminchisescu, C.
\newblock Deep learning of graph matching.
\newblock In \emph{2018 {IEEE} Conference on Computer Vision and Pattern Recognition, {CVPR} 2018, Salt Lake City, UT, USA, June 18-22, 2018}, pp.\  2684--2693. Computer Vision Foundation / {IEEE} Computer Society, 2018.
\newblock \doi{10.1109/CVPR.2018.00284}.

\bibitem[Zhang et~al.(2020)Zhang, Liu, Zhou, and Zhang]{zhang2020hybrid}
Zhang, H., Liu, F., Zhou, Y., and Zhang, Z.
\newblock A hybrid method integrating an elite genetic algorithm with tabu search for the quadratic assignment problem.
\newblock \emph{Information Sciences}, 539:\penalty0 347--374, 2020.

\bibitem[Zhang et~al.(2023)Zhang, Liu, Li, Zhen, Yuan, Li, and Yan]{DBLP:journals/ijon/ZhangLLZYLY23}
Zhang, J., Liu, C., Li, X., Zhen, H., Yuan, M., Li, Y., and Yan, J.
\newblock A survey for solving mixed integer programming via machine learning.
\newblock \emph{Neurocomputing}, 519:\penalty0 205--217, 2023.
\newblock \doi{10.1016/J.NEUCOM.2022.11.024}.

\end{thebibliography}
\bibliographystyle{icml2024}


\newpage
\appendix
\onecolumn
\textbf{APPENDIX}
\section{Network details}
\label{apend_net}
Our SAWT model consists of an encoder and a decoder. At first, we initialize the facility nodes with vectors randomly sampled from a one-hot vector pool with a dimension of $N_{init} = 128$. Details are shown below in Python code.

\begin{lstlisting}[language=Python]
    f_init_emb = torch.zeros(size=(batch_size, node_cnt, N_{init})).to(device)
    # shape: (batch, node, embedding)
    
    seed_cnt = 100
    rand = torch.rand(batch_size, seed_cnt)
    # Create a random one-hot vector pool for each instance within a batch
    
    batch_rand_perm = rand.argsort(dim=1)
    rand_idx = batch_rand_perm[:, :node_cnt]
    # Collect one-hot vector index for facility nodes, ``node_cnt" refers to the problem size
    
    
    b_idx = torch.arange(batch_size)[:, None].expand(batch_size, node_cnt)
    n_idx = torch.arange(node_cnt)[None, :].expand(batch_size, node_cnt)
    f_init_emb[b_idx, n_idx, rand_idx] = 1
    # Create the one-hot vector
\end{lstlisting}


In our model, facility and location nodes are first processed through a linear function, producing outputs of dimension $d_{emb}$. We utilize a single-mixed-score Transformer with two modules, each featuring eight heads, alongside a three-layer Graph Convolution Network (GCN), each layer having a hidden dimension of $d_{hidden}$. The facility and location node representations are then merged and processed through $L$ modules of the SAWT Transformer, maintaining the eight-head configuration. The SAWT Transformer's output feeds into both policy and value decoders. The policy decoder consists of two Multi-Layer Perceptrons (MLPs): the first, $MLP_1$, with three layers sized $(2 \times d_{hidden}, d_{hidden})$, $(d_{hidden}, d_{hidden})$, and $(d_{hidden}, 1)$, and the second, $MLP_2$, also with three layers but sized $(3 \times d_{hidden}, d_{hidden})$, $(d_{hidden}, d_{hidden})$, and $(d_{hidden}, 1)$. These MLPs aim to select a pair of positions for swap operations to generate a new solution state $\sigma^{'}$. Meanwhile, the value decoder evaluates the current and optimal solutions $(\sigma^{'}, \sigma^{\ast})$ to produce a scalar value.

The training parameters in our model are updated using the loss function below:
\begin{equation}
    \mathcal{L} = -J(\theta) + 0.99^{epoch} \beta H(\theta) + \zeta \mathcal{L}(\phi).
\end{equation}

The definitions of $J(\theta)$, $H(\theta)$, and $\mathcal{L}(\phi)$ are provided in Subsection \ref{RL}. The hyper-parameters $\beta$ and $\zeta$ are set to 0.005 and 0.5, respectively, for all experiments. The term $0.99^{epoch}$ is used to shift the model's focus towards exploitation in the later training stages. Our policy training closely follows the REINFORCE algorithm, with bootstrapping steps $T = 8$ and a discount factor $\gamma = 0.99$ as detailed in Subsection \ref{RL}. Training includes a total of 400 search steps. The Adam optimizer is employed for gradient updates. Table \ref{hyper} below summarizes the hyper-parameter settings:

\begin{table}[ht]
\caption{Hyperparameter values for each task.}
\centering
\begin{sc}
\begin{tabular}{c|cccc}
\hline
     Hyperparameter& QAP10 & QAP20 & QAP50 & QAP100  \\
     \hline
      $d_{emb}$& 64& 64& 64& 64\\
      $d_{hidden}$ &64& 64& 64& 64 \\
      $L$ &3&2&3&3\\
      $lr$ &$10^{-3}$&$10^{-3}$&$10^{-3}$&$10^{-3}$\\
      \hline
\end{tabular}
\label{hyper}
\end{sc}
\end{table}

\section{Theoretical insight}
\label{theory}

The goal of the Quadratic Assignment Problem (QAP) is to minimize the following equation:
\begin{eqnarray}
&\mathop{\min}\limits_{\mathbf{X}} & trace (\mathbf{F} \cdot \mathbf{X} \cdot \mathbf{D} \cdot \mathbf{X}^{T}), \\ & s.t. & \mathbf{X} \cdot \mathbf{1} = \mathbf{1}, \\ && \mathbf{X}^{T} \cdot \mathbf{1} = \mathbf{1}, \\ && \mathbf{X} \in\{0,1\} ^{n \times n}.
\end{eqnarray}


The gradient of the objective function in any direction represented by matrix $\mathbf{V}$ is given by:

\begin{equation}
\renewcommand{\arraystretch}{1.7}
\setlength{\arraycolsep}{2pt}
\begin{array}{ll} &\langle D_X trace(\mathbf{F} \cdot \mathbf{X} \cdot \mathbf{D} \cdot \mathbf{X}^T) , \mathbf{V} \rangle \\
= &\langle D_X trace( \mathbf{X} \cdot \mathbf{D}^T \cdot \mathbf{X}^T \cdot \mathbf{F}^T), \mathbf{V}\rangle \vspace{2ex}\\ = & lim_{t \rightarrow 0} \frac{trace( (\mathbf{X} + t\mathbf{V})\cdot \mathbf{D}^T \cdot (\mathbf{X} + t\mathbf{V})^T \cdot \mathbf{F}^T ) - trace( \mathbf{X} \cdot \mathbf{D}^T \cdot \mathbf{X}^T \cdot \mathbf{F}^T)}{t} \vspace{2ex}\\  =& lim_{t \rightarrow 0} \frac{ t^2 \cdot trace( \mathbf{V} \cdot \mathbf{D}^T\cdot \mathbf{V}^T \cdot \mathbf{F}^T) + t \cdot (trace( \mathbf{X} \cdot \mathbf{D}^T \cdot \mathbf{V}^T \cdot \mathbf{F}^T) + trace( \mathbf{V} \cdot \mathbf{D}^T \cdot \mathbf{X}^T\cdot \mathbf{F}^T))}{t} \vspace{2ex}\\  = &trace( \mathbf{X} \cdot \mathbf{D}^T \cdot \mathbf{V}^T \cdot \mathbf{F}^T) + trace(\mathbf{V} \cdot \mathbf{D}^T \cdot \mathbf{X}^T\cdot \mathbf{F}^T) \vspace{2ex}\\  =& trace(\mathbf{F}^T \cdot \mathbf{X} \cdot \mathbf{D}^T \cdot \mathbf{V}^T ) + trace( \mathbf{F} \cdot \mathbf{X}\cdot \mathbf{D}\cdot \mathbf{V}^T) \vspace{2ex}\\  = &\langle \mathbf{F}^T \cdot \mathbf{X} \cdot \mathbf{D}^T, \mathbf{V} \rangle + \langle \mathbf{F} \cdot \mathbf{X}\cdot \mathbf{D}, \mathbf{V} \rangle \vspace{2ex}\\  =& \langle \mathbf{F}^T \cdot \mathbf{X} \cdot \mathbf{D}^T+\mathbf{F} \cdot \mathbf{X}\cdot \mathbf{D}, \mathbf{V} \rangle. \\ 
\end{array} 
\end{equation}


Following our derivation, we find $D_X trace(\mathbf{F} \cdot \mathbf{X} \cdot \mathbf{D} \cdot \mathbf{X}^T) = \mathbf{F}^T \cdot \mathbf{X} \cdot \mathbf{D}^T+\mathbf{F} \cdot \mathbf{X}\cdot \mathbf{D}$. In our self-generated datasets, where $\mathbf{F} = \mathbf{F}^T$ and $\mathbf{D} = \mathbf{D}^T$, this simplifies to: $D_X trace(\mathbf{F} \cdot \mathbf{X} \cdot \mathbf{D} \cdot \mathbf{X}^T) = 2 \cdot \mathbf{F} \cdot \mathbf{X}\cdot \mathbf{D}$. This expression matches matrix $\mathbf{M}$ from Equation \ref{SAWT-Att} in our main paper, which is used in the Solution Aware Transformer. We propose that the Solution Aware Transformer's design, incorporating the objective gradient into the attention layer, aids in pattern learning for QAPs. By applying Equation \ref{update rl} from our main paper, the reinforcement learning update gradient can be steered correctly, thereby enhancing performance.

\section{QAPLIB benckmarks}
\label{apend_QAPLIB}


The QAPLIB\footnote{\url{https://coral.ise.lehigh.edu/data-sets/qaplib/}} \cite{burkard1997qaplib} includes 134 real-world QAP instances across 15 categories, such as planning and hospital facility layout \cite{elshafei1977hospital}, with problem sizes ranging from 12 to 256. Instances are named based on a specific rule: the prefix denotes the problem category (usually the author's name), followed by a number indicating the problem size. If multiple problems share the same size, a letter starting from $a$ is added for distinction. For example, see Equation \ref{qap instanc}.

\begin{equation}
\label{qap instanc}
\underset{\text { author name }}{\underline{\text { bur }}} -\underset{\text { problem size }}{\underline{26}} - \underset{\text{index}}{\underline{a}}.
\end{equation}

\section{Baseline details}
\label{apend_base}


We provide detailed information about the methods used in our experiments. Gurobi and TS are exact solvers that perform exhaustive searches for solutions. SM, RRWM, IPFP, and Sinkhorn-JA are heuristics that efficiently handle Lawler's QAPs as shown in Equation \ref{lawler QAP}:
\begin{equation}
\label{lawler QAP}
\renewcommand{\arraystretch}{1.5}
\setlength{\arraycolsep}{2pt}
\begin{array}{ll}
    \mathop{\min}\limits_{\mathbf{X}}& \displaystyle vec(\mathbf{X}) \cdot \mathbf{K} \cdot vec(\mathbf{X})^{T}, \\
\text{s.t.} & \displaystyle \mathbf{X} \cdot \mathbf{1} = \mathbf{1}, \mathbf{X}^{T} \cdot \mathbf{1} = \mathbf{1}, \\
& \mathbf{X} \in\{0,1\} ^{n \times n} .
\end{array}
\end{equation}


The association graph $\mathbf{K} \in \mathbb{R}^{n^2 \times n^2}$ is central to various optimization problems. MatNet is an L2C method designed for solving asymmetric TSPs and flexible flow shop problems. In contrast, Costa, Sui, Wu, Dact, and NeuOpt are L2I methods that address VRPs. Additionally, NGM and RGM are learning-based approaches for solving Lawler's QAPs.

\textbf{Gurobi} \cite{gurobi2021gurobi} is a leading optimization solver, particularly effective in solving combinatorial optimization problems (COPs). It primarily uses the Branch-and-Bound method to perform exhaustive solution searches.

\textbf{TS} \cite{zhang2020hybrid} 
This approach integrates a genetic method into tabu-search algorithms. In each iteration, the genetic method selects an elite solution for the tabu list using crossover and mutation processes. The tabu-search algorithm follows the pseudocode from \cite{zhang2020hybrid}.

\textbf{SM} \cite{leordeanu2005spectral} 
considers matching problems as identifying graph clusters in the association graph using traditional spectral techniques. The concept is that nodes to be matched should cluster together.

\textbf{RRWM} \cite{cho2010reweighted} 
adopts a reweighted random-walk algorithm on the association graph to solve matching problems.

\textbf{IPFP} \cite{leordeanu2009integer} 
refines the solution iteratively through integer projection.

\textbf{SK-JA} \cite{kushinsky2019sinkhorn} 
utilizes the widely recognized Johnson–Adams (JA) relaxation, a linear program relaxation of the Quadratic Assignment Problem, to solve instances from QAPLIB.

\textbf{MatNet$^{\ast}$} \cite{DBLP:conf/nips/KwonCYPPG21} 
is modified from MatNet, an L2C method that directly encodes matrix inputs. MatNet employs a dual-mixed attention model to integrate the matrix input into the attention mechanism.

\textbf{Costa} \cite{d2020learning} 
The model employs an RNN to capture the sequence in TSP solutions. Its decoder utilizes a pointer network to select two locations for a 2-opt operation.

\textbf{Sui} \cite{sui2021learning} 
It employs a 3-opt heuristic for TSPs, utilizing a link selection network to create fresh connections among the selected trio of locations.

\textbf{Wu} \cite{wu2021learning} 
The VRPs employ Transformers, a versatile tool widely used in deep learning applications \cite{tan2024vision,hu2023dacdetr,hu2023suppressing,hu2022switch}. Wu et al. \cite{wu2021learning} pioneered the integration of Transformers into VRPs. Positional embeddings within the encoder capture the linear structure of VRP solutions. Subsequently, the decoder computes location similarities, selecting the highest scores for 2-opt, swap, or insert operations.

\textbf{Dact} \cite{ma2021learning} 
Introducing an innovative dual-aspect collaborative Transformer for encoding locations, it incorporates a unique cyclic positional encoding. This feature effectively captures the circularity and symmetry inherent in Vehicle Routing Problems (VRPs).

\textbf{NeuOpt} \cite{ma2023learning}  
demonstrates prowess in executing flexible k-opt exchanges using a specialized action factorization method, comprising three key moves: S-move, I-move, and E-move. It utilizes a distinct recurrent dual-stream decoder tailored for decoding these exchanges. Moreover, NeuOpt incorporates a novel Guided Infeasible Region Exploration strategy to help the model navigate out of infeasible regions in Capacitated Vehicle Routing Problems (CVRPs).

\textbf{NGM} \cite{wang2021neural} 
is the inaugural learning-based approach to tackle Lawler's QAPs. Initially crafted for resolving graph matching (GM) problems, NGM's applications extend to diverse fields like video segmentation \cite{jiang2023video}. In addressing QAPs, NGM transforms them into a constrained vertex classification challenge on the association graph, denoted as $\mathbf{K}$. This graph emerges from the amalgamation of the flow and distance matrices, $\mathbf{F}$ and $\mathbf{D}$, respectively, as per Equation \ref{associate}.

\begin{equation}
\label{associate}
\mathbf{K} = \mathbf{F} \otimes \mathbf{D} = 
\begin{bmatrix}
f_{11}\mathbf{D} & \cdots & f_{1n}\mathbf{D} \\
\vdots & \ddots & \vdots \\
f_{m1}\mathbf{D} & \cdots & f_{mn}\mathbf{D}
\end{bmatrix}.
\end{equation}
The association graph is acquired through vertex classification by an embedding network, then normalized with Sinkhorn method, and trained end-to-end using cross-entropy loss.

\textbf{RGM} \cite{liu2022revocable} 
is the inaugural RL technique to conquer the QAPLIB benchmark, capable of training sans ground truth. A revocable action framework enhances the agent's adaptability for intricate constrained vertex classification tasks on association graphs.

\section{Details of the MatNet$^{\ast}$}
\label{apend_mat}

\begin{figure}[ht]
    \centering
    \includegraphics[width=0.7\linewidth]{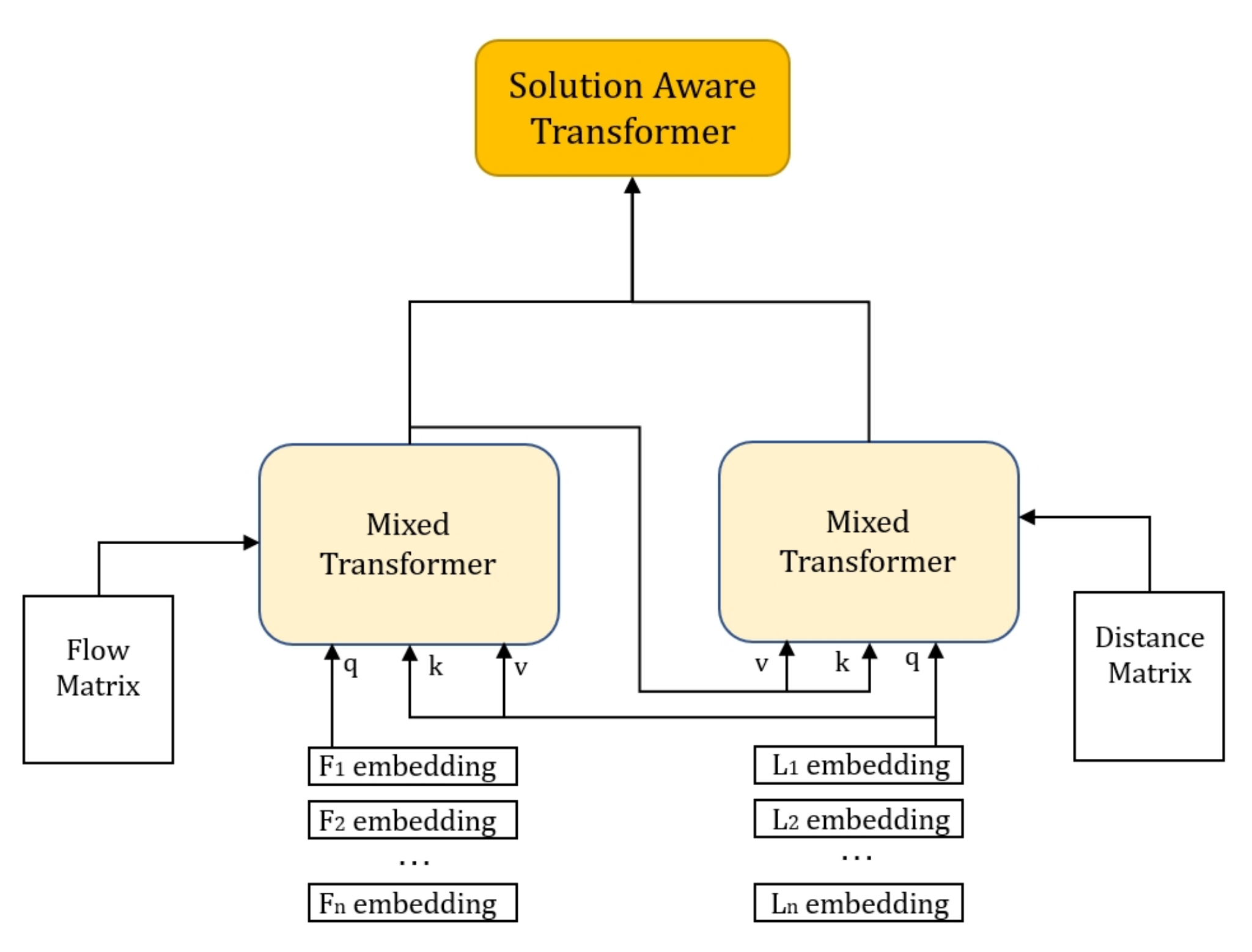}
    \caption{Structure of MatNet$^{\ast}$.}
    \label{matnet}
\end{figure}

We have repurposed MatNet, originally an L2C method for asymmetric TSPs and flexible flow shop problems, for our QAP tasks as an L2I method. Our adaptation, MatNet$^{\ast}$, employs a dual-attention mechanism for encoding facility and location nodes (Figure \ref{matnet}). Following the approach in \cite{DBLP:conf/nips/KwonCYPPG21}, facility nodes start with 128-dimensional random initial embeddings, while location nodes begin with zero vectors of the same dimension. These, along with flow and distance matrices, undergo processing through three modules of a mixed-score transformer before input into our SAWT model for swap operations. However, this approach is computationally intensive due to the use of two transformers for encoding.

\section{More experimental results}
\label{apend_exp}

\subsection{Experiments with additional metrics}


Two additional metrics are provided, standard deviation (Std) and winning rate, to assess SAWT. Following the methodology of ~\cite{d2020learning}, Std is computed across 512 testing instances for each QAP task. The winning rate is determined by comparing solution instances between SAWT and TS${5K}$. Results are summarized in Tables \ref{metric std} and \ref{win}. Key findings from Table \ref{metric std} include: (1) SAWT exhibits superior performance, evidenced by lower median and reduced standard deviation. (2) RGM, utilizing the L2C method, shows higher standard deviation compared to L2I methods, indicating the robustness of L2I approaches. (3) Heuristic methods struggle with both median and standard deviation, highlighting challenges in achieving robust, optimal QAP solutions. Analysis from Table \ref{win} reveals that while SAWT achieves better average objective values compared to TS${5K}$, its winning rate against TS${5K}$ remains moderate. This suggests the existence of challenging instances for SAWT relative to TS${5K}$.

\begin{table}[H]
\centering
\caption{Standard deviation experiments on QAP tasks.}
\begin{sc}
\begin{tabular}{l|c|c|c}
\hline \multirow{2}{*}{methods}  & QAP 20 & QAP50& QAP100 \\
\cline{2-4} & Mean $\pm$ Std & Mean $\pm$ Std & Mean $\pm$ Std   \\
\hline
SM & 70.07 $\pm$ 7.93&446.62 $\pm$ 22.85&1800.75 $\pm$ 61.21 \\
RRWM &73.74 $\pm$ 8.56&457.40 $\pm$ 23.14& 1823.76 $\pm$ 72.32\\
IPFP &75.52 $\pm$ 8.54&479.44 $\pm$ 23.66& 1911.90 $\pm$ 72.54 \\
\hline
Costa &57.91 $\pm$ 6.98&404.28 $\pm$ 22.32 &1720.62 $\pm$ 55.52\\
DACT &60.47 $\pm$ 7.04 & 418.77 $\pm$ 21.92& 1743.82 $\pm$ 56.11\\
RGM & 55.96 $\pm$ 7.51& --&--\\
\hline
SAWT &54.63 $\pm$ \textbf{6.68}&379.96 $\pm$ \textbf{20.10}&1617.23 $\pm$ \textbf{52.23}\\
\hline
\end{tabular}
\end{sc}
\label{metric std}
\end{table}

\begin{table}[H]
\centering
\caption{Winning rate experiments on QAP tasks.}
\begin{sc}
\begin{tabular}{l|c|c|c|c}
\hline
Winning rate  &QAP10 & QAP 20 & QAP50 & QAP100 \\
\hline
SAWT & 100.00\% &72.26\% & 67.42\% &10.15\%\\
\hline
\end{tabular}
\end{sc}
\label{win}
\end{table}

\subsection{Additional experiments on QAP tasks}

\begin{figure}[t]
  \centering
  \subfigure[Gaps on 256 validation instances for 400 steps over training epochs.]{
\begin{minipage}[t]{0.5\linewidth}
\centering
\includegraphics[width=3.2in]{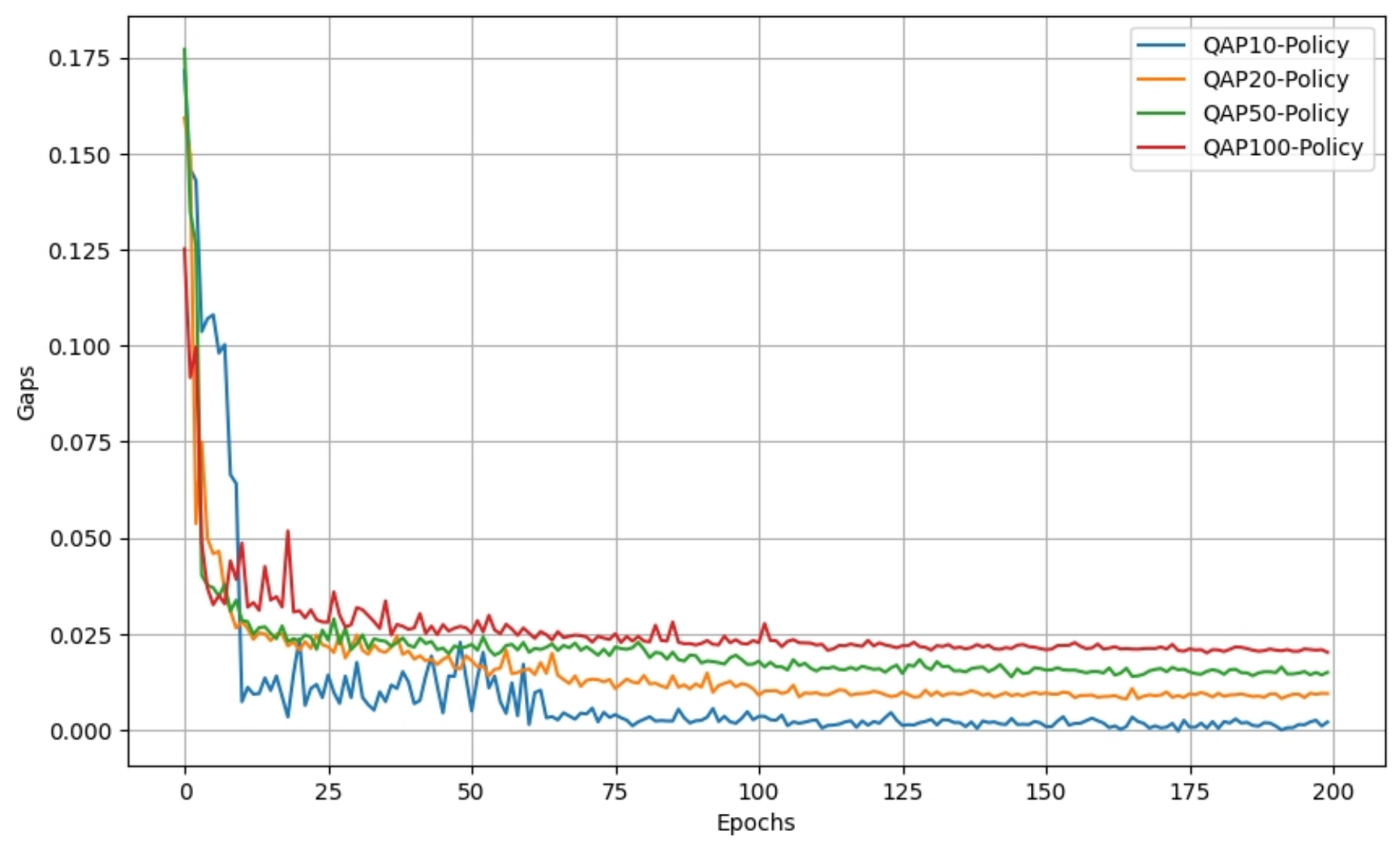}
\end{minipage}%
}%
\subfigure[Gaps on 256 testing instances over 10000 search steps.]{
\begin{minipage}[t]{0.5\linewidth}
\centering
\includegraphics[width=3.2in]{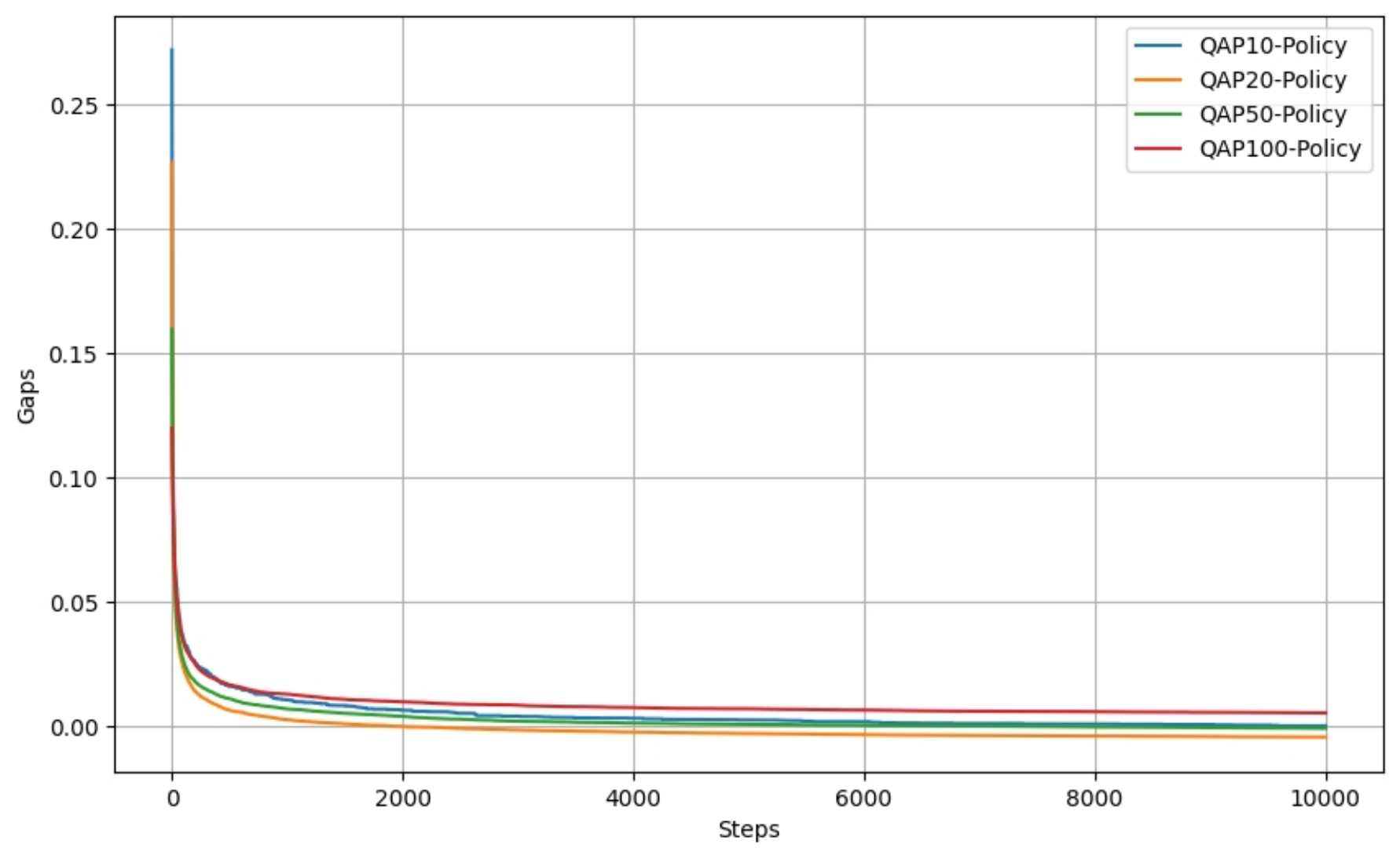}
\end{minipage}%
}%
\caption{Model behavior on training and testing period. }
\label{train_val}
\end{figure}

In Figure \ref{train_val}, we illustrate the optimal gap evolution for policies trained on QAP10, QAP20, QAP50, and QAP100, across training and testing phases. The gap, averaged over 256 validation instances across 400 search steps during training, and similarly over 256 testing instances across 10,000 search steps during testing, showcases a progressive narrowing throughout training epochs. Notably, Figure \ref{train_val} (a) indicates larger instance sizes correlating with wider gaps, indicating increased difficulty levels. Additionally, Figure \ref{train_val} (b) illustrates the swift gap minimization in the initial learning phases, transitioning to a more nuanced fine-tuning process as learning advances.

\subsection{Additional experiments on TSP tasks}

To test SAWT on solving MILPs, we conducted experiments on TSP tasks. Following ~\cite{ma2023learning}, Our SAWT model was trained on self-generated datasets TSP20, TSP50, and TSP100. During inference, we searched for solutions over 10,000 steps and averaged the objective values. From Table \ref{TSP}, Our method proves comparable to Dact and surpasses AM and Costa, as indicated by the results. Although SAT performed less favorably than NeuOpt, it suggests the promising potential for SAT in TSP solving, considering SAT lacks modules tailored for TSPs. Considering that the compared baselines face extreme difficulties in solving QAP tasks as shown in Table \ref{Main results}, we highlight the importance of our work.

\begin{table}[H]
\centering
\caption{Experiments on TSP using SAWT. We train SAWT on TSP training data, and test it on the test data using 10000 search steps.}
\begin{sc}
\begin{tabular}{l|cc|cc|cc}
\hline \multirow{2}{*}{methods}   & \multicolumn{2}{|c|}{TSP20} & \multicolumn{2}{|c|}{TSP50} & \multicolumn{2}{|c}{TSP100} \\
\cline{2-7} & Mean$\downarrow$ & gap$\downarrow$  & mean$\downarrow$ & gap$\downarrow$  & mean$\downarrow$ & gap$\downarrow$  \\
\hline
AM &3.828 &0.01\% &5.699 &0.05\% &7.811 &0.60\% \\
COSTA  &3.827&0.00\%&5.703&0.12\%&7.824&0.77\%\\
Dact &3.827&0.00\%&5.696&0.00\%&7.772&0.10\%\\
Neuopt &\textbf{3.827}&\textbf{0.00\%}&\textbf{5.696}&\textbf{0.00\%}&\textbf{7.766}&\textbf{0.02\%}\\
\hline
SAWT &3.827  &0.00\% &5.698 &0.04\%&7.782&0.22\%\\
\hline
\end{tabular}
\end{sc}
\label{TSP}
\end{table}

\subsection{Inference time of SAWT}

Table \ref{Inference time} demonstrates the runtimes for the SAWT encoder, decoder, and full model. Two key observations emerge: (1) The SAWT encoder's runtime is nearly triple that of the decoder during inference, owing to its more intricate design. (2) The total runtime deviates from the sum of the encoder and decoder runtimes due to our model's action sampling via the CPU and possible delays between the CPU and GPU.

\begin{table*}[t]
\caption{Inference time of SAWT's encoder, decoder and full model.}
\centering
\scriptsize
\begin{sc}
\begin{tabular}{l|ccc|ccc|ccc}
\hline \multirow{2}{*}{method}  & \multicolumn{3}{|c|}{QAP20} & \multicolumn{3}{|c|}{QAP50} & \multicolumn{3}{|c}{QAP100} \\
\cline{2-10} & Encoder & Decoder  &  Full model & Encoder & Decoder  &  Full model   & Encoder & Decoder  &  Full model  \\
\hline
SAWT$\{10\text{k}\}$ &75.48s  &26.84s  &152s	&77.85s& 27.41s  &160s	&229.14s& 85.52s & 452s
\\
\hline
\end{tabular}
\end{sc}
\label{Inference time}
\end{table*}

\subsection{Full experiments on QAPLIB}
\label{qaplib_append}
In this section, we present our generalization experiments to QAPLIB by directly applying our trained policy on QAP50. The results are shown in Table \ref{tab:my_label}, and results are copied from NGM's project \footnote{https://thinklab.sjtu.edu.cn/project/NGM/index.html}. There are four observations: (1) Our method is capable of solving all instances on a single GPU equipped with 12GB of memory. In contrast, as documented in \cite{wang2021neural}, NGM requires more than 48GB of memory to solve a single instance when the size exceeds 150. This underscores the efficiency of our distinct encoding strategy. (2) Our methods surpass all the heuristics and learning-based methods in most of the instances. This proves the strong generalization ability of our SAWT model. (3) For large instances with $\ast$, our SAWT outperforms almost all the instances which further demonstrates the ability to solve large instances of our model. (4) For instances within the ``Chr," ``Ste," and ``Esc" categories, the observed gaps are considerably substantial—for instance, 201.2\% for ``Chr20a" and 206.2\% for ``Esc128"—highlighting the limitations of our model.

\subsection{Robustness analysis on QAPLIB}

From Table \ref{tab:my_label}, SAWT demonstrates a promising ability to tackle large-scale QAPs. For instance, our model can solve the ``Tho150" using only 1 GB of GPU memory, a task at which RGM fails and NGM requires over 36GB of GPU memory. Instances marked with an asterisk ($\ast$) indicate RGM's failure, typically on instances larger than 70. SAT excels in 86\% (18 out of 21) of these large instances compared to NGM, a testament to its robustness on real-world large-scale datasets.

To delve deeper to understand the results in Table \ref{QAPLIB} and Table \ref{tab:my_label}, we've integrated the flow matrix's density statistics in Table \ref{robust QAPLIB}, as it is one of the primary differences among datasets. In Section \ref{instance generation}, we generate our flow matrix with a density value of $p=0.7$, resulting in a relatively dense matrix. There are four observations: (1) Our model, SAWT, exhibits varying performance across different datasets with diverse flow matrix densities. Specifically, it struggles with ``Chr", ``Kra", ``Ste", and ``Tho", which have low flow matrix densities, indicating sensitivity to matrix density. Despite this, SAWT outperforms NGM  and competes favorably with RGM, highlighting its robust QAP pattern learning ability. (2) Interestingly, SAWT performs well on ``Esc" (29.6\%) and ``Scr" (38.8\%), even with their low matrix densities. A closer look reveals that ``Esc" has local density, which SAT perceives as a dense matrix, thereby improving performance. For instance, the flow matrix of ``Esc16A" features a 10 $\times$ 10 dense matrix in the upper left corner, with only diagonal elements equal to zero. The remainder of the matrix's elements are zero. ``Scr" instances are simpler QAPs, as indicated by Gurobi's quick optimal solutions (solved ``Scr12", ``Scr15" within $1s$ and ``Scr20" within $5min$), which may explain SAWT's fairly good performance. (3) SAWT generalizes well to datasets with high flow matrix densities, irrespective of different non-zero element distributions, illustrating its capacity to learn common QAP patterns. (4) A notable observation from Table \ref{tab:my_label} is SAWT's better performance on ``TaiA" (88.8\%) compared to ``TaiB" (49.1\%), further underscoring its sensitivity to flow matrix density.

\begin{table}[t]
\centering
\tiny
\caption{Data distribution of QAPLIB.}
\begin{sc}
\begin{tabular}{l|c|c|c|c|c|c|c|c|c|c|c|c|c|c}
\hline
   & bur&chr&esc&had&kra&lipa&nug&rou&scr&sko&ste&tai(a/b)&tho&wil \\
\hline
Flow density & 77.6\% &15.2\% &29.6\% &91.6\%&36.6\%&89.2\%&62.5\%&90.2 \%& 38.8\% & 68.3\% &26.5 \% &88.8\%/49.1\% &48.2 \%& 87.9\%  \\
\hline
\end{tabular}
\end{sc}
\label{robust QAPLIB}
\end{table}

\begin{table}[ht]
\centering
\begin{sc}
\begin{tabular}{c|c|ccc|c|cc}
\hline
instance & upper bound &sm &rrwm & sinkhorn-ja&ngm&SAWT& gap\\
\hline
bur26a & 5426670&6533340	&6663181	&5688893	&5684628&\textbf{5587628} &2.9\%\\
bur26b &3817852&	4690772&	4741283&	4053243	&4063246& \textbf{3903418}& 2.3\%\\
bur26c &5426795	&6537412&	6474996&	5639665	&5638641& \textbf{5592521} & 3.1\%\\
bur26d &3821225&	4649645&	4678974	&3985052&	3994147&\textbf{3951321}& 3.4\%\\
bur26e &5386879	&6711029&	6619788&	5539241	&5666202& \textbf{5570198}& 3.4\%\\
bur26f &3782044&	4723824&	4814298&	3979071&	3954977& \textbf{3868905}& 2.2\%\\
bur26g &10117172&	12168111&	12336830&	10624776	&10855165& \textbf{10437749}& 3.1\%\\
bur26h &7098658&	8753694&	8772077&	7453329	&7670546& \textbf{7344127}& 3.4\%\\
chr12a &9552&	50732	&43624	&\textbf{9552}	&27556&21046& 120.4\%\\
chr12b& 9742	&46386	&73860&	\textbf{9742}&	29396&19522& 100.3\%\\
chr12c &11156	&57404&	50130&	\textbf{11156}&	34344&22966& 105.1\%\\
chr15a &9896&	77094	&90870	&\textbf{11616}	&50272&19342& 95.4\%\\
chr15b &7990	&77430	&115556&	\textbf{7990}&	52082&20014& 152.2\%\\
chr15c& 9504	&64198	&70738	&\textbf{9504}	&38568&20592& 116.6\%\\
chr18a& 11098	&94806	&115328	&\textbf{11948}	&83026&25808& 132.5\%\\
chr18b& 1534	&4054	&3852	&\textbf{2690}	&4810&3522& 130.3\%\\
chr20a &2192	&11154	&13970&	\textbf{4624}	&10728&6615& 201.2\%\\
chr20b& 2298	&9664	&14168&	\textbf{3400}	&9962&5998& 161.0\%\\
chr20c &14142	&112406&	195572&	\textbf{40464}	&115128&41146& 190.9\%\\
chr22a &6156	&16732	&15892	&\textbf{9258}	&16410&7085& 15.1\%\\
chr22b &6194	&13294	&13658	&\textbf{6634}	&15876&7488& 20.8\%\\
chr25a &3796	&21526&	32060	&5152 &	18950&\textbf{4126}& 8.6\%\\
els19$^{\ast}$ &17212548&	33807116&	74662642&	\textbf{18041490}	&34880280& 21825512&26.8\%\\
esc16a &68	&98	&80 &100	&88& \textbf{68} & 0.0\%\\
esc16b &292	&318	&\textbf{294}	&304	&308&296 &1.3\%\\
esc16c &160	&276	&204	&266	&184&\textbf{162} & 1.2\%\\
esc16d &16	&48	&44	&58	&40&\textbf{16} & 0.0\%\\
esc16e &28	&52	&50	&44	&48&\textbf{32} &14.2\%\\
esc16f &0	&0	&0	&0	&0& \textbf{0} & 0.0\%\\
esc16g &26	&44	&52	&52	&50&\textbf{26} & 0.0\%\\
esc16h &996	&1292	& \textbf{1002}	&1282	&1036&1030& 3.4\%\\
esc16i &14	&54	&28	&36	&26&\textbf{18} & 28.5\%\\
esc16j &8	&22	&18	&18&16& \textbf{8} & 0.0\%\\
esc32a& 130&	426&	240	&456	&428& \textbf{212} & 63.0\\
esc32b &168	&460&	400	&416&	424& \textbf{252} & 50.0\%\\
esc32c &642	&770	&650	&886	&844& \textbf{650} &1.2\%\\
esc32d &200	&360&	\textbf{224}	&356&	288 & 246 & 23.0\%\\
esc32e&2	&68	&6	&46&	42 & \textbf{2} &0.0\%\\
esc32g& 6	&36	&10	&46	&28 & \textbf{10} & 63.3\%\\
esc32h &438	&602	&\textbf{506}	&--	&592 & 530 & 21.0\%\\
esc64a &116	&254	&\textbf{124}	&276 &250& 190 & 63.7\%\\
esc128$^{\ast}$ &64	&202&	\textbf{78}	&--	&238& 196 & 206.2\%\\
had12 &1652&	1894	&2090	&--	&1790& \textbf{1696} & 2.6\%\\
had14 &2724	&3310	&3494	&2916&	2922& \textbf{2826} & 3.7\%\\
had16 &3720	&4390	&4646	&3978	&4150& \textbf{3854} & 3.6\%\\
had18 &5358&	6172	&6540	&5736	&5780& \textbf{5574} & 4.0\%\\
had20 &6922&	8154	&8550	&7464	&\textbf{7334}& 7372 & 6.5\%\\
kra30a &88900	&148690	&136830&	125290	&\textbf{114410}& 117436&32.1\% \\
kra30b &91420	&150760	&141550	&126980	&118130& \textbf{117110}& 28.1\%\\
kra32  &88700	&145310	&148730	&128120	&121340 & \textbf{119036}& 34.2\%\\
\hline
\end{tabular}
\end{sc}
\end{table}

\begin{table}[ht]
\centering
\begin{sc}
\begin{tabular}{c|c|ccc|c|cc}
\hline
instance & upper bound &sm &rrwm & sinkhorn-ja&ngm&SAWT& gap\\
\hline
lipa20a &3683	&3956&	3940&	3683&	3929&  \textbf{3683} & 0.0\%\\
lipa20b  &27076&	36502&	38236&	27076&	33907& \textbf{27076} & 0.0\%\\
lipa30a  &13178	&13861&	13786&	13178	&13841& \textbf{13178} & 0.0\%\\
lipa30b  &151426&	198434&	201775&	151426&	192356&\textbf{151426} & 0.0\% \\
lipa40a  &31538	&32736	&32686	&31538	&32666& \textbf{31538} & 0.0\%\\
lipa40b  &476581&	628272&	647295&	476581&	616656&\textbf{476581} & 0.0\%\\
lipa50a  &62093	&64070&	64162	&62642	&64100&\textbf{62093} &0.0\%\\
lipa50b  &1210244&	1589128&	1591109&	\textbf{1210244}&	1543264&1241710 &2.6\%\\
lipa60a & 107218&	109861&	110468&	108456	&110094& \textbf{107218} &0.0\%\\
lipa60b  &2520135	&3303961&	3300291&	\textbf{2520135}&	3269504& 2578098 & 2.3\%\\
lipa70a$^{\ast}$  &169755	&173649	&173569	&\textbf{172504}	&173862& 173153 & 2.0\%\\
lipa70b$^{\ast}$  & 4603200	&6055613	&6063182&	4603200&	5978316& \textbf{4603200} & 0.0\%\\
lipa80a$^{\ast}$ & 253195	&258345	&258608	&\textbf{257395}	&258402& 257778 & 1.8\%\\
lipa80b$^{\ast}$  &7763962	&10231797&	10223697&	7763962&	10173155& \textbf{7763962} & 0.0\%\\
lipa90a$^{\ast}$  &360630	&367384	&367370	&366649	&367193& \textbf{366573} & 1.6\%\\
lipa90b$^{\ast}$  &12490441	&16291267&	16514577&	12490441&	16194745& \textbf{12490441} & 0.0\%	\\
nug12 &578	&886&	1038&	682&	720& \textbf{602} & 4.1\% \\
nug14 &1014	&1450&	1720&	--&	1210 & \textbf{1068} & 5.3\%\\
nug15  &1150&	1668&	2004&	1448&	1482& \textbf{1234} & 7.3\%\\
nug16a &1610&	2224&	2626&	1940&	1836& \textbf{1738} & 7.9\%\\
nug16b &1240&	1862&	2192&	1492&	1580& \textbf{1340} & 8.0\%\\
nug17 &1732	&2452&	2934&	2010&	2004 & \textbf{1906} & 10.0\%\\
nug18 &1930	&2688&	3188&	2192&	2312 & \textbf{2128} & 10.2\%\\
nug20 &2570	&3450&	4174&	3254&	2936 & \textbf{2826} & 9.9\%\\
nug21  &2438&	3702&	4228&	3064&	2916& \textbf{2680} & 10.0\%\\
nug22  &3596&	5896&	6382&	3988&	4616&	\textbf{3956} & 10.1\%\\
nug24 &3488	&4928	&5720&	4424&	4234& \textbf{3844} & 10.5\%\\
nug25 &3744	&5332	&5712&	4302&	4420& \textbf{4126} & 10.4\%\\
nug27  &5234&	7802&	8626&	6244&	6332& \textbf{5810} & 11.1\%\\
nug28 &5166	&7418	&8324&	6298&	6128 & \textbf{5734} & 11.0\%\\
nug30 &6124	&8956	&10034&	7242&	7608& \textbf{6926} & 13.1\% \\
rou12 &235528	&325404&	377168&	276446&	321082& \textbf{254842} & 8.2\% \\
rou15  &354210&	489350	&546526	&\textbf{390810}	&469592& 393882 & 11.2\%\\
rou20 &725522&	950018	&1010554&	823298&	897348& \textbf{820566} & 13.1\%\\
scr12 &31410&	71392&	95134	&45334&	44400& \textbf{34614} & 10.2\%\\
scr15 &51140&	104308&	101714&	74632	&81344& \textbf{65102} & 27.3\%\\
scr20 &110030&	263058&	350528&	171260	&182882& \textbf{162846} & 48.0\% \\
sko42 &15812	&20770&	23612&	19058&	20192& \textbf{18660} & 18.7\%\\
sko49 &23386	&29616&	34548&	\textbf{27160}&	28712& 27292 & 16.7\%\\
sko56 &34458	&44594&	49650&	40954&	42182& \textbf{40662} & 18.0\%\\
sko64 &48498	&60878&	65540&	\textbf{55738}&	60368& 57034 & 17.6\%\\
sko72$^{\ast}$ &66256	&82156&	89264&	76332&	79716&\textbf{76508} & 15.4\%\\
sko81$^{\ast}$ &90998	&112838&	118372&	105246&	107588& \textbf{104726} &15.0\% \\
sko90$^{\ast}$ &115534	&140840	&148784&	133818&	137402& \textbf{132270} &14.4\%\\
sko100a$^{\ast}$ &152002	&185738&	184854&	176626&	180972& \textbf{173798} & 14.3\%\\
sko100b$^{\ast}$ &153890	&185366&	189502&	177398&	180774 & \textbf{176576} & 14.7\%\\
sko100c$^{\ast}$ &147862	&178710&	188756&	169566&	175740& \textbf{169290} & 14.4\%\\
sko100d$^{\ast}$ &149576	&181328&	186086&	\textbf{170648}&	175096& 170964	& 14.3\%\\
sko100e$^{\ast}$ &149150	&180062&	192342&	171656&	176010& \textbf{170858} & 14.5\%	\\
sko100f$^{\ast}$ &149036&	177518&	189284&	171296&	173552& \textbf{170404} & 14.3\%\\
\hline
\end{tabular}
\end{sc}
\end{table}

\begin{table}[ht]
\centering
\small
\begin{sc}
\begin{tabular}{c|c|ccc|c|cc}
\hline
instance & upper bound &sm &rrwm & sinkhorn-ja&ngm&SAWT& gap\\
\hline
ste36a &9526&	30030&	33294&	17938&	16648& \textbf{13972} &46.7\%\\
ste36b &15852&	176526&	193046&	47616&	43248& \textbf{42956} & 170.9\% \\
ste36c &8239110&	24530792&	28908062&	14212212&	\textbf{12988352}& 13418384	& 62.9\%\\
tai12a &224416&	318032&	392004&	\textbf{245012}&	259014&249102 &11.1\% \\
tai12b &39464925&	96190153&	124497790&	81727424&	65138752&\textbf{46173962} & 16.9\% \\
tai15a  &388214&	514304&	571952&	471272&	467812& \textbf{390155} & 0.5\%\\
tai15b &51765268&	702925159&	702292926&	\textbf{52585356}&	495479040&52718274 &1.7\% \\
tai17a &491812&	669712&	738566&	598716&	630644& \textbf{559682} & 13.8\%\\
tai20a &703482&	976236&	1012228&	849082&	896518&\textbf{794934} & 12.9\% \\
tai20b &122455319&	394836310&	602903767&	220470588&	237607744& \textbf{139599064} & 14.0\%\\
tai25a &1167256&	1485502&	1536172&	\textbf{1341104}	&1393248&1371526 &17.5\%\\
tai25b &344355646&	764920942&	1253946482&	798113083&	730775168& \textbf{512056846} & 48.7\%\\
tai30a &1818146&	2210304&	2305048&	2072218&	2065706	& \textbf{1930871} & 6.2\%\\
tai30b &637117113&	1008164383&	1766978330&	1114514832&	1359600384&\textbf{828252247} & 30.0\%\\
tai35a &2422002&	3030184	&3100748&	2820060	&2886132& \textbf{2669068} & 10.2\%\\
tai35b &283315445&	454981851&	574511546&	446783959&	455718176& \textbf{360377246} & 27.2\%\\
tai40a &3139370&	3825396&	3985684&	\textbf{3547918}	&3610604& 3619378 & 15.2\%\\
tai40b &637250948&	1165811212&	1423772477&	1019672934&	1053339520& \textbf{ 900201547} & 41.2\%\\
tai50a &4938796	&6078426&	6203546&	\textbf{5569952}	&5891066& 5590717& 13.2\%\\
tai50b &458821517&	796553600&	790688128&	696556852&	764856128&\textbf{632256050} & 37.8\%\\
tai60a &7205962&	8614998&	8731620&	8243624&	8596094& \textbf{8230248} &14.2\%\\
tai60b &608215054&	1089964672&	1279537664&	978843717&	994559424& \textbf{801450909} &31.7\%	\\
tai64c &1855928&	5893540&	6363888&	3189566&	5703540& \textbf{2080496} &12.1\%\\
tai80a$^{\ast}$ &13499184&	15665790&	16069786&	15352662&	15648708& \textbf{15279496} & 13.1\%\\
tai80b$^{\ast}$ &818415043&	1338090880&	1410723456&	1215586531&	1275809408& \textbf{1108782694} & 35.4\%\\
tai100a$^{\ast}$& 21052466&	24176962&	24446982&	23787764&	24077728& \textbf{23576714} & 11.9\%\\
tai100b$^{\ast}$ &1185996137&	1990209280&	2192130048&	\textbf{1589275900}&	1853681152& 1677867703 &41.4\%\\
tai150b$^{\ast}$ &498896643&	662657408&	755505920&	--&	653429440& \textbf{627949553} & 25.8\%\\
tho30 &149936&	230828&	267194&	202844&	187062&\textbf{184722} & 23.2\%\\
tho40 &240516&	375154&	440146&	314070&	313026& \textbf{304012} & 26.4\%\\
tho150$^{\ast}$ &8133398&	10000616&	10689758&	9508422&	9702946& \textbf{9595620} & 17.9\%\\
wil50 &48816&	56588&	60420&	54030&	55390& \textbf{52700} & 8.1\%\\
wil100$^{\ast}$ &273038&	305030&	307258&	\textbf{292118}&	295418& 294568& 7.8\%\\
\hline

\end{tabular}
\end{sc}
\caption{The $\ast$ in the Table means the instances RGM fails to solve due to the large size of the instances. We do not add the RGM method for the reason that the author does not reveal the independent results on each instance. The instances tested both on RGM and SAWT results in a mean average of $35.8\%$ and $26.8\%$ which demonstrates the effectiveness of our SAWT. Some instances like ``Chr" and ``Ste" perform poorly on the SAWT, showing the limitation of our model's generalization ability.  }
\label{tab:my_label}
\end{table}

\end{document}